\documentclass[3p,12pt]{elsarticle}

\usepackage{graphicx}
\usepackage{amsmath,amssymb} % define this before the line numbering.
\usepackage{bm}
\usepackage{footmisc}
\usepackage{algorithmic}
\usepackage{algorithm}
\usepackage{multirow}
\usepackage{fancyref}
\usepackage{footmisc}
\usepackage{array}
\usepackage{wrapfig}
\usepackage{xcolor}
\usepackage{array,graphicx,subfigure}

\usepackage{multirow}

\usepackage{color}

\usepackage{amsfonts}

\usepackage{amsmath,amssymb}

\usepackage{mathrsfs}

\usepackage[justification=centering]{caption}

\usepackage{url}

\newcommand{\PreserveBackslash}[1]{\let\temp=\\#1\let\\=\temp}
\newcolumntype{C}[1]{>{\PreserveBackslash\centering}p{#1}}
\newcolumntype{R}[1]{>{\PreserveBackslash\raggedleft}p{#1}}
\newcolumntype{L}[1]{>{\PreserveBackslash\raggedright}p{#1}}

\journal{Neurocomputing}

\begin{document}

\begin{frontmatter}



%\title{Cross-modal Subspace Learning for Inter-category and Intra-category Sketch-based Image Retrieval: A Comparative Study}
\title{Cross-modal Subspace Learning for Fine-grained Sketch-based Image Retrieval}


\author{Peng~Xu$^1$, Qiyue~Yin$^2$, Yongye~Huang$^1$, Yi-Zhe~Song$^3$, Zhanyu~Ma$^1$$^*$ \footnotetext{$*$~Corresponding author.}, Liang~Wang$^2$,\\ Tao Xiang$^3$, W.~Bastiaan~Kleijn$^4$, Jun~Guo$^1$}

\address{$^1$Pattern Recognition and Intelligent System Laboratory, \\
        Beijing University of Posts and Telecommunications, Beijing, China \\
         \{peng.xu, yongye, mazhanyu, guojun\}@bupt.edu.cn \\
        $^2$National Lab of Pattern Recognition, \\
        Institute of Automation, Chinese Academy of Sciences, Beijing, China \\
        \{qyyin, wangliang\}@nlpr.ia.ac.cn \\
        $^3$SketchX Lab, School of Electronic Engineering and Computer Science, \\
        Queen Mary University of London, London, UK \\
        \{yizhe.song, t.xiang\}@qmul.ac.uk \\
        $^4$Communications and Signal Processing Group, Victoria University of Wellington, New Zealand\\
         bastiaan.kleijn@ecs.vuw.ac.nz}

\begin{abstract}
Sketch-based image retrieval (SBIR) is challenging due to the inherent domain-gap between sketch and photo.
Compared with pixel-perfect depictions of photos, sketches are iconic renderings of the real world with highly abstract.
Therefore, matching sketch and photo directly using low-level visual clues are unsufficient, since a common low-level subspace that traverses semantically across the two modalities is non-trivial to establish. Most existing SBIR studies do not directly tackle this cross-modal problem.
This naturally motivates us to explore the effectiveness of cross-modal retrieval methods in SBIR, which have been applied in the image-text matching successfully.
In this paper, we introduce and compare a series of state-of-the-art cross-modal subspace learning methods and benchmark them on two recently released fine-grained SBIR datasets. Through thorough examination of the experimental results, we have demonstrated that the subspace learning can effectively model the sketch-photo domain-gap.
In addition we draw a few key insights to drive future research.
\end{abstract}


\end{frontmatter}

% \linenumbers


\section{Introduction}

Sketch-based image retrieval (SBIR) has drawn increasingly more attention in the past decade, especially with the prevalence of touchscreens.
There exist many annotated datasets~\cite{qianyu2016sketch,eitz2010evaluation,hu2013performance,ouyang2014cross,keli2016fgsbir,Sangkloy2016TheSD} and methods tackling all aspects of the problem~\cite{BMVC.28.115,qianyu2016sketch,keli2016fgsbir,Yu2016ijcv,xupeng2016Instance,jifei2016Multi}.
The vibrancy of the SBIR area also promoted the development of other related research problems,
such as sketch recognition~\cite{zhang2015recognition,li2015free}, sketch synthesis~\cite{Gao20081921,Xiao2010840,li2016free}, sketch-based $3$D retrieval~\cite{su2015multi},
and sketch segmentation~\cite{qi2015making}. From a technical perspective, SBIR is traditionally cast into a classification task, with most prior work evaluating the retrieval performance at category-level~\cite{li2015free,cao2010mindfinder,cao2011edgel,qi2015im2sketch,wang2015sketch}.
More recently, fine-grained variants of SBIR ~\cite{BMVC.28.115,qianyu2016sketch} requires retrieval to be conducted within single object categories. With this more constrained ranking setting of the problem, researchers no longer carry out similarity matching based only on low-level and hand-designed visual features~\cite{eitz2010evaluation,hu2013performance,shrivastava2011data}, but begin to devolve into high-level and partial information for sketch and photo matching, e.g., local stroke ordering~\cite{Yu2016ijcv,yu2015bmvcsketch}, and part-level attributes~\cite{ouyang2014cross,keli2016fgsbir,xupeng2016Instance}.

\begin{figure*}[htb]
\centering
\includegraphics[width=0.90\textwidth]{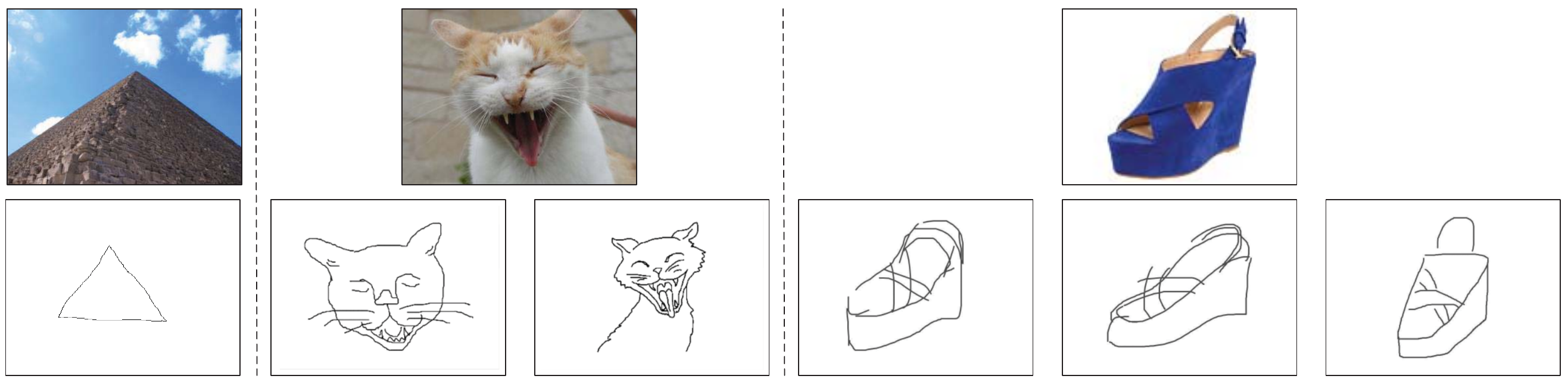}%
\caption{Illustration of abstraction and diversity of sketches.}
\label{domain-gap}
\end{figure*}

Despite great strides made, most prior work ignores the cross-modal gap that inherently exists between sketch and photo, treating images as edgemaps (semi-sketches)~\cite{BMVC.28.115,qi2015making,qianyu2016sketch,jifei2016Multi}.
This assumption works well when the retrieval system is presented with good quality sketches that are close to contour tracings of intended objects, but would not work well with free-hand sketches where sketches are much more abstract and do not offer close resemblance natural objects. However, effectively solving the sketch-photo cross-modal gap is non-trivial:
(\romannumeral1)~Sketch can only capture limited shape and contour information.
 It utilizes coarse lines to describe key features of an object at an abstract and semantic level.
 As shown in Fig.~\ref{domain-gap}, a pyramid can be denoted as a triangle in sketch form.
(\romannumeral2)~Different people have different observations, past experiences, drawing styles, and drawing skill~\cite{keli2016fgsbir,Sangkloy2016TheSD}.
Fig.~\ref{domain-gap} shows us that sketches drawn by different persons for the same cat or shoe may be highly diverse. This naturally motivates us to apply cross-modal matching methods to tackle the SBIR problem.
However, to the best of our knowledge, all previous cross-modal work~\cite{rasiwasia2010new,sharma2012generalized,zhen2012co,masci2014multimodal,Xu2016191,Yao2016250} are designed to address the image-text modal gap (e.g., Wikipedia image-text dataset~\cite{rasiwasia2010new}, Pascal VOC dataset~\cite{DBLP:journals/pami/HwangG12}, NUS-WIDE~\cite{nus-wide-civr09}, LabelMe~\cite{Oliva2001}).
Therefore, making their general applicableness to SBIR remains unclear.

The main approaches behind existing image-text cross-modal research can be roughly categorized into pair-wise modeling~\cite{zhen2012co,quadrianto2011learning,zhai2013heterogeneous,Wang2015imagetext},
ranking~\cite{David2008Discriminative,Jason2011Wsabie,Huang2016125}, mapping~\cite{hardoon2004canonical,wang2013learning,Wang2015Twostep}, and graph embedding~\cite{sharma2012generalized,Wang2015Twostep,Li2016501,DBLP:journals/pami/WangHWWT16}.
In particular, probabilistic models~\cite{putthividhy2010topic,jia2011learning}, metric learning approaches~\cite{wu2010learning,mignon2012cmml,zhu2013linear,liu2013hypergraph},
and subspace learning methods~\cite{hardoon2004canonical,kim2007discriminative} have been proven to be effective across a number of datasets.
Probabilistic approaches learn the multi-modal correlation by modeling the joint multi-modal data distribution~\cite{putthividhy2010topic}.
Metric learning methods learn and compute appropriate distance metrics between different modalities~\cite{wu2010learning}.
Subspace learning constructs the common subspace and map multi-modal data into it to conduct cross-modal matching~\cite{wang2013learning}.
Among these cross-modal techniques, cross-modal subspace learning methods have achieved state-of-the-art results in recent years~\cite{rasiwasia2010new,wang2013learning,DBLP:journals/pami/WangHWWT16,costa2014role,sharma2011bypassing}, which have borrowed much inspiration from the conventional subspace approaches~\cite{liu2010robust,liu2011latent,liu2013robust,6909596,chang2014convex,chang2015convex,chang2016compound}. For a comprehensive survey, please refer to~\cite{kaiye2016comparative, liu2010cross}.

In this paper, we focus on analyzing the interaction and relationship between sketch and photo in the cross-modal setting.
The main contributions of this paper are two-fold:
\begin{itemize}
  \item We present and compare a series of state-of-the-art cross-modal subspace learning methods,
        and benchmark them on
        two recently released fine-grained SBIR datasets~\cite{qianyu2016sketch,keli2016fgsbir}.
  \item We conduct detailed comparative analysis towards the general applicability of cross-modal techniques on matching sketches and photos.
\end{itemize}

The remaining parts of this paper are organized as follows.
%Section $2$ discusses the theoretical advantages of the subspace learning methods comparing with other cross-modal approaches.
Section $2$ briefly presents some state-of-the-art cross-modal subspace learning methods and the corresponding characteristics.
In Section $3$, we report and analyze their experimental performances for the SBIR task.
Potential future research insights for SBIR are discussed in Section $4$.
Finally, conclusions are drawn in Section $5$.

A preliminary version of this work has been presented in~\cite{xupeng2016Cross}.
The main extensions are:
\begin{itemize}
  \item We add three cross-modal subspace learning methods (CDFE~\cite{lin2006inter}, CCA-3V~\cite{gong2014multi}, JFSSL~\cite{DBLP:journals/pami/WangHWWT16}) for intensive comparisons.
  \item Extensive experiments are performed on one extra recently released fine-grained SBIR dataset (i.e., the chair SBIR dataset).
   %\item     Throughout the experimental validation, different metrics are adopted for different level SBIR task. \textcolor{blue}{did you use new metrics? if not, delete this point, if yes, say which}
    \item    We simultaneously evaluate the performances of these methods for SBIR tasks on both subcategory-level and instance-level.
\end{itemize}

\section{Cross-modal Subspace Learning}
In this section, we will briefly survey some state-of-the-art cross-modal subspace learning methods designed for image and text.
All these methods will share the same notation.
Suppose that sample matrices ${\bf X}_a = [{\bf x}^a_1, {\bf x}^a_2, \cdots, {\bf x}^a_n] \in \mathbb{R}^{d_a \times n}$ and ${\bf X}_b = [{\bf x}^b_1, {\bf x}^b_2, \cdots, {\bf x}^b_n] \in \mathbb{R}^{d_b \times n}$ are extracted from modality $a$ and modality $b$, respectively.
These multi-modal samples can be categorized into $c$ classes. Samples and the corresponding class labels are denoted as $\{ {\bf x}^a_i, c^a_i \}^n_{i=1}$~and~$\{ {\bf x}^b_i, c^b_i \}^n_{i=1}$,
where each pair $\{ {\bf x}^a_i, {\bf x}^b_i \}~(1 \leqslant i \leqslant n )$ represents the same object or content belonging to the same class.
${\bf Y} = [ {\bf y}_1, {\bf y}_2, \cdots, {\bf y}_n ]^T \in \mathbb{R}^{n \times c}$ denotes the class label matrix for the multi-modal data.
The transform for the $i$-th sample of modality $a$ is denoted as ${\bf x}^a_i \rightarrow  {\bf f}^a_i$.
Similarly, the transform for the $i$-th sample of modality $b$ is denoted as ${\bf x}^b_i \rightarrow {\bf f}^b_i$.
Throughout this paper, vectors and matrices are denoted as straight bold lower-case ${\bf x}$ and upper-case ${\bf X}$, respectively.

These cross-modal subspace learning methods have the common workflow of learning a projection matrix for each modality to project the data from different modalities into a common comparable subspace in the training phase.
In the test phase, the test data samples from one modality will be taken as the query set to retrieve matched samples from the other modality.
In this paper, ${\bf W}_a$ and ${\bf W}_b$ denote the projection matrices for modality $a$ and modality $b$, respectively.

\subsection{Canonical Correlation Analysis (CCA)}

CCA~\cite{rasiwasia2010new,hardoon2004canonical,kim2007discriminative} is an effective multivariate statistical analysis approach, which is analogous to principal component analysis (PCA)~\cite{jolliffe2002principal}. It was originally designed for data correlation modeling and dimension reduction. Recently, CCA has been applied widely in multi-modal data fusion and cross-media retrieval~\cite{rasiwasia2010new,costa2014role,gong2014multi,7410823}.
CCA has become one of the most popular unsupervised cross-modal subspace learning methods due to its generalization capability.

CCA learns a set of canonical component pairs for ${\bf X}_a$ and ${\bf X}_b$, i.e., directions~${\bf w}_a \in \mathbb{R}^{d_a}$~and~${\bf w}_b \in \mathbb{R}^{d_b}$~along which the multi-modal data is maximally correlated~\cite{rasiwasia2010new} as
\begin{equation}
\label{equ:cca}
\max \limits_{{\bf w}_a\neq0, {\bf w}_b\neq0} \frac{{\bf w}_a^T{\bf \Sigma}_{ab}{\bf w}_b}{\sqrt{{\bf w}_a^T{\bf \Sigma}_{aa}{\bf w}_a}\sqrt{{\bf w}_b^T{\bf \Sigma}_{bb}{\bf w}_b}}~,
\end{equation}
where ${\bf \Sigma}_{aa}$~and~${\bf \Sigma}_{bb}$~denote the empirical covariance matrices for~$a$~modality and~$b$~modality, respectively.
${\bf \Sigma}_{ab}={\bf \Sigma}_{ba}^T$~represents the cross-covariance matrix between different modalities.
By repeatedly solving~\eqref{equ:cca}, we can obtain a series of canonical component pairs.
We can choose the first $d$ canonical component pairs $\{{\bf w}_a, {\bf w}_b\}_i~(1 \leqslant i \leqslant d)$ for projecting ${\bf X}_a$ and ${\bf X}_b$ into two $d$ dimensional subspaces.
Here, $d$ is a hyper-parameter.
This optimization objective of~\eqref{equ:cca} can be solved as a generalized eigenvalue problem (GEV)~\cite{ramsay2006functional}.

\subsection{Partial Least Squares (PLS)}

PLS~\cite{rosipal2006overview} can linearly map multi-modal data into a linear subspace that preserves the data correlation.
It can be adopted to solve the cross-modal matching in many multi-modal scenarios.
PLS has been effectively applied in face recognition and multi-media retrieval
with different motivations~\cite{sharma2011bypassing,baek2004face,dhanjal2009efficient,li2009maximizing,vstruc2009gabor,schwartz2010robust,chen2012continuum}.

PLS models ${\bf X}_a$ and ${\bf X}_b$ such that~\cite{sharma2011bypassing}
\begin{equation}
\begin{split}
{\bf X}_a^T = {\bf TP}^T+{\bf E}\\
{\bf X}_b^T = {\bf UQ}^T+{\bf F}\\
{\bf U} = {\bf TD}+{\bf H}~.\\
\end{split}
\end{equation}
${\bf T} \in \mathbb{R}^{n\times d}$ and ${\bf U} \in \mathbb{R}^{n\times d}$ contain the $d$ extracted PLS scores or latent projections.
${\bf P} \in \mathbb{R}^{p\times d}$~and ${\bf Q} \in \mathbb{R}^{q\times d}$~are the matrices of loadings and ${\bf E} \in \mathbb{R}^{n \times p}$,
~${\bf F} \in \mathbb{R}^{n \times q}$, and~${\bf H} \in \mathbb{R}^{n \times d}$~are the residual matrices. ${\bf D} \in \mathbb{R}^{d \times d}$~
is a diagonal matrix describing the latent scores of ${\bf X}^T_a$~and~${\bf X}^T_b$.

PLS learns the basis vectors ${\bf w}_a$ and ${\bf w}_b$~such that the covariance
between the score vectors ${\bf t}$ and ${\bf u}$~(rows of ${\bf T}$ and ${\bf U}$) is maximized as
\begin{equation}
\begin{aligned}
\max([\mathrm{cov}({\bf t}, {\bf u})]^2) &=\max_{{\bf w}_a, {\bf w}_b}([\mathrm{cov}({\bf X}^T_a{\bf w}_a, {\bf X}^T_b{\bf w}_b)]^2)\\
\mathrm{s.t.}~\Vert{\bf w}_a\Vert &=\Vert{\bf w}_b\Vert=1~.
\end{aligned}
\end{equation}

\subsection{Generalized Multi-view Analysis (GMA)}
%Generalized Multi-view Linear Discriminant Analysis (GMLDA), and Generalized Multi-view Margin Fisher Analysis (GMMFA)
GMA~\cite{sharma2012generalized} is a special multi-view framework,
which can be solved efficiently as a generalized eigenvalue problem.
As we will show in this section, many popular supervised and unsupervised feature extraction techniques can be derived based on GMA.

The constrained objective is
\begin{equation}
\begin{split}
\max_{{\bf w}_a, {\bf w}_b} {\bf w}_a^T {\bf A}_a {\bf w}_a &+ \mu {\bf w}_b^T {\bf A}_b {\bf w}_b
+ \beta {\bf w}_a^T {\bf X}_a {\bf X}_b^T {\bf w}_b   \\
\mathrm{s.t.}~{\bf w}_a^T {\bf B}_a {\bf w}_a &+\alpha {\bf w}_b^T {\bf B}_b {\bf w}_b = 1~,\\
\end{split}
\label{equ:gma}
\end{equation}
where, ${\bf w}_a$~and~${\bf w}_b$~denote the projection directions.
The positive terms $\mu$, $\beta$, and $\alpha$ are hyper-parameters controlling the balance among the objectives.
${\bf A}_i~(i = a, b)$ is the between-class variance matrix while ${\bf B}_i~(i = a, b)$ is the within-class covariance matrix.
Sharma et al.~\cite{sharma2012generalized} have illustrated that if we substitute $\{{\bf A}_i,~{\bf B}_i,~{\bf X}_ i\}_{i = a, b}$ in~\eqref{equ:gma} with particular expressions, we obtain the corresponding objective functions of different methods.

\subsubsection{Bilinear Model (BLM)}
In~\eqref{equ:gma}, setting ${\bf A}_i = {\bf X}_i {\bf X}_i^T / n$, ${\bf B}_i = {\bf I}$, and we obtain BLM under the proposed GMA framework.

\subsubsection{Generalized Multi-view Linear Discriminant Analysis (GMLDA)}
We can set ${\bf A}_i = {\bf S}^B_i$, ${\bf B}_i = {\bf S}^W_i$,
where ${\bf S}^W/{\bf S}^B$ are the within/between-class scatter matrices.
Here, ${\bf X}_i$ is substituted by its class mean matrix.

\subsubsection{Generalized Multi-view Margin Fisher Analysis (GMMFA)}
Based on GMA framework, the expression for the multi-view version of MFA is complex.
It utilizes the graph construction to restrict the projected data.
More details can be found in~\cite{sharma2012generalized}.

\subsection{Common Discriminant Feature Extraction (CDFE)}

Lin and Tang~\cite{lin2006inter} used the~\emph{empirical separability} and the~\emph{local consistency} to propose the CDFE method for subspace learning.
The~\emph{empirical separability} ensures the intra-class compactness and the inter-class dispersion, which are measured respectively as follows~\cite{lin2006inter}
\begin{equation}
\label{equ:cdfe-intraclass}
J_{intraclass} = \frac{1}{N_1} \sum _{i = 1}^{n} \sum_{j:c^b_j = c^a_i} \| {\bf f}^a_i - {\bf f}^b_j \|^2~,
\end{equation}
\begin{equation}
\label{equ:cdfe-interclass}
J_{interclass} = \frac{1}{N_2} \sum _{i = 1}^{n} \sum_{j:c^b_j \neq c^a_i} \| {\bf f}^a_i - {\bf f}^b_j \|^2~,
\end{equation}
where $N_1$ and $N_2$ are the quantities of sample pairs from the same class and the different classes, respectively.

\begin{figure}[htb]
\centering
\includegraphics[width=0.40\textwidth]{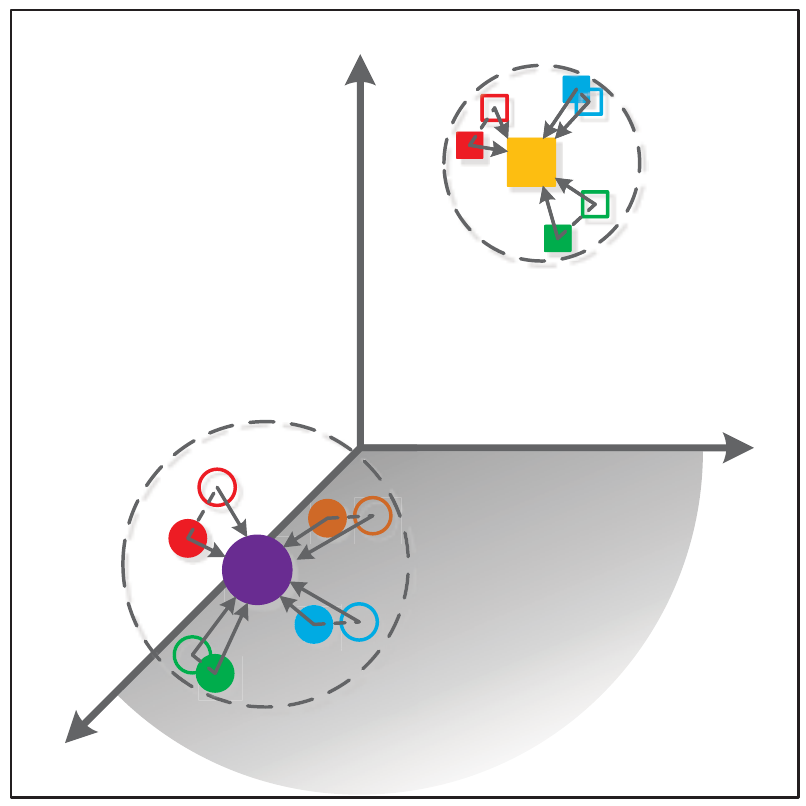}%
\caption{Diagram of CDFE~\emph{empirical separability}.
The smaller solid and hollow figures represent the data from the different modalities. The smaller figures with the same shape
and different colors represent different samples belonging to the same class. The bigger figures with different shapes indicate the corresponding domain-independent semantic labels.
The short dashed straight lines visualize the pair-wise relationships.
The short bold arrows link multi-modal sample pairs to their semantic labels.
The short dashed straight lines visualize the pair-wise relationships.
%The short bold arrows link multi-modal sample pairs to their semantic labels.
The bold black dashed circles describe the~\emph{empirical separability}.
The  intra-class compactness can be intuitively understood as compressing the two circles.
The inter-class dispersion can be described as keeping them far from each other.
}
\label{graph-construction}
\end{figure}

As shown in Fig.~\ref{graph-construction}, the~\emph{empirical separability} can be defined as:
\begin{equation}
\label{equ:cdfe-empirical}
J_{e} = J_{intraclass} - \alpha J_{interclass}~,
\end{equation}
where $\alpha$ is a hyper-parameter for trade-off.
To prevent the overfitting,~\emph{local consistency} can be used to regularize the~\emph{empirical separability}.
The objective function of CDFE can be formulated as follows:
\begin{equation}
\label{equ:cdfe-objective}
J_{CDFE} = J_{e} +\beta J_{l}~,
\end{equation}
where $\beta$ is a hyper-parameter to adjust the trade-off between these two objectives.
Here, $J_{l}$ represents the~\emph{local consistency} objective.
More details can be found in~\cite{lin2006inter}.

\begin{figure*}[htb]
	\centering
	\subfigure[Subspace learned by CCA-2V]{
		\label{fig:cca-3v-a}
		\includegraphics[width=0.40\textwidth]{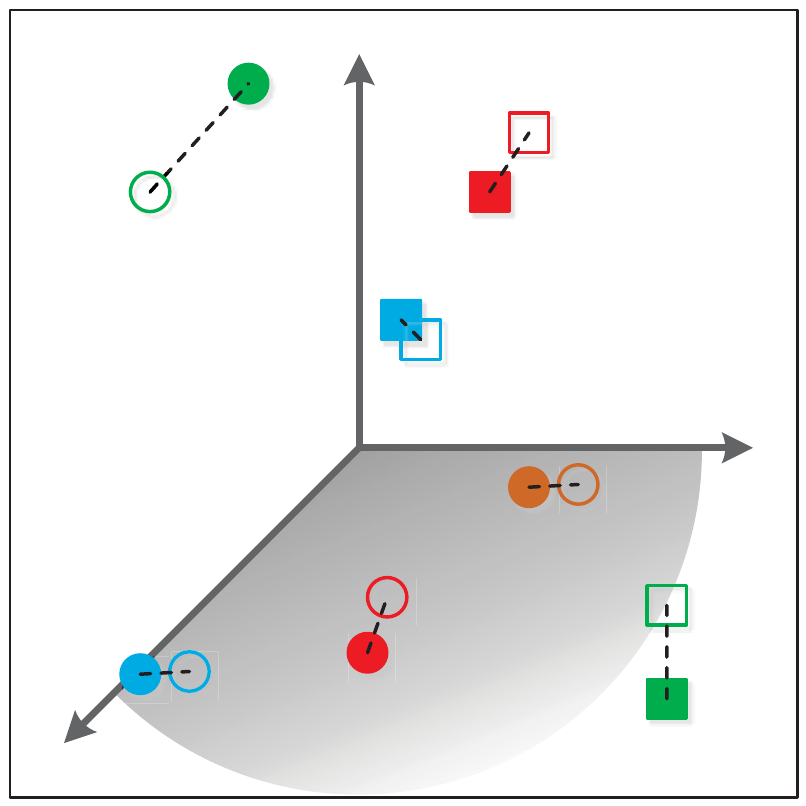}}
	\hspace{5mm}	
	\subfigure[Subspace learned by CCA-3V]{
		\label{fig:cca-3v-b}
		\includegraphics[width=0.40\textwidth]{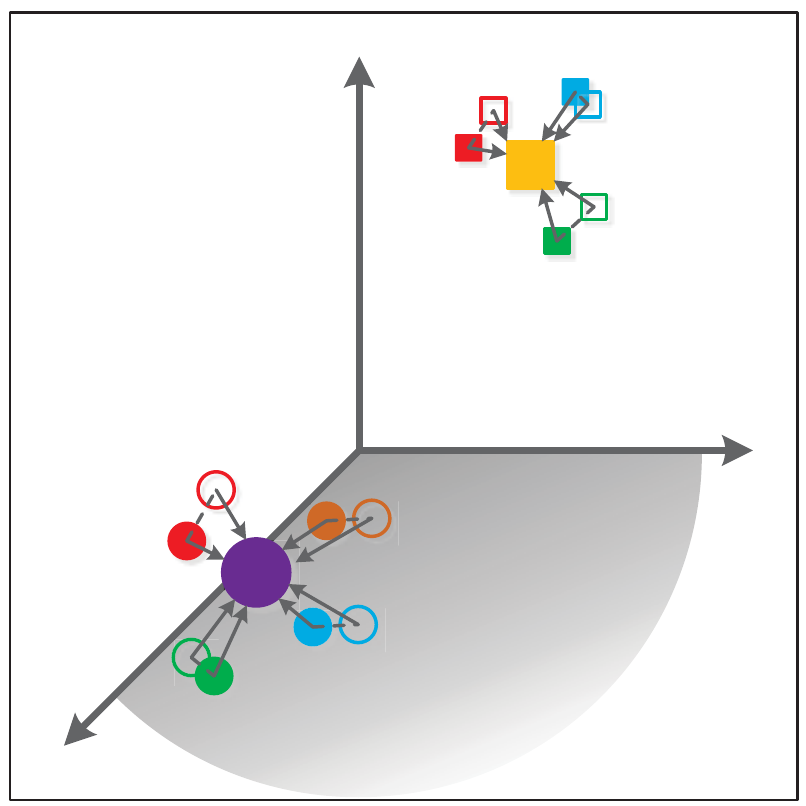}}
	\caption{The differences between CCA-2V and CCA-3V.
$(a)$ Traditional two-view CCA pair-wise maximizes the correlation.
$(b)$ Three-view CCA incorporates semantic classes as a third view.
The smaller solid and hollow figures represent the data from the different modalities.
The smaller figures with the same shape and different colors represent different samples belonging to the same class.
The bigger figures with different shapes indicate the corresponding domain-independent semantic labels.
The short dashed straight lines visualize the pair-wise relationships.
The short bold arrows link multi-modal sample pairs to their semantic labels.
}
	\label{fig:cca-3v}
\end{figure*}

\subsection{Three-view Canonical Correlation Analysis (CCA-3V)}

The objective function of CCA-3V has three terms~\cite{gong2014multi}:
\begin{equation}
\label{equ:cca-3v}
\begin{split}
\min \limits_{{\bf W}_a, {\bf W}_b, {\bf W}_c} \Vert {\bf X}^T_a {\bf W}_a - {\bf X}^T_b {\bf W}_b \Vert_F^2 + ~~~~~~~~~~~~ \\
\Vert {\bf X}^T_a {\bf W}_a - {\bf X}^T_c {\bf W}_c \Vert_F^2 +
\Vert {\bf X}^T_b {\bf W}_b - {\bf X}^T_c {\bf W}_c \Vert_F^2~.
\end{split}
\end{equation}
Obviously, the latent correlation among three views or three modalities can be captured by optimising this function.
Moreover, for the cross-modal matching, some high-level semantic information can be utilized as the third view~\cite{gong2014multi}.
If we put the ground-truth labels into its third view, it becomes a supervised method.
As shown in Fig.~\ref{fig:cca-3v}, comparing with the conventional CCA, CCA-3V constructs a semantic embedding subspace to improve the performance.
CCA-3V aligns the corresponding multi-modal sample pairs by not only referring to the data distribution but also following the guidance of the high-level semantics.
Multi-modal samples belonging to the same semantic cluster are forced to be close to each other.

\subsection{Learning Coupled Feature Spaces for Cross-modal Matching (LCFS)}

Many earlier studies have demonstrated two properties:
\begin{itemize}
  \item $l_{21}$-norm has good performances in feature selection~\cite{gu2011joint,he20122,NIPS2010_3988}.
  \item The trace norm~\cite{angst2011generalized,fornasier2011low,grave2011trace,harchaoui2012large} can model the correlation of the design matrix or prior knowledge as the low-rank solution.
\end{itemize}

Integrating the properties of the $l_{21}$-norm and the trace norm,
Wang et al.~\cite{wang2013learning} proposed a model of the following form
\begin{equation}
\label{equ:lcfs}
\begin{split}
\min_{{\bf W}_a, {\bf W}_b} \frac{1}{2}(\Vert{\bf X}_a^T {\bf W}_a - {\bf Y}\Vert_F^2 + \Vert{\bf X}_b^T {\bf W}_b - {\bf Y}\Vert_F^2)\\
+\lambda_1(\Vert{\bf W}_a\Vert_{21}+\Vert{\bf W}_b\Vert_{21})     + \lambda_2\Vert[{\bf X}_a^T {\bf W}_a~{\bf X}_b^T {\bf W}_b]\Vert_* ~,
\end{split}
\end{equation}
where ${\bf W}_a$ and ${\bf W}_b$ are the projection matrices for the coupled $a$~modality and~$b$ modality, respectively.
The first term is a coupled linear regression, which is used to learn two projection matrices for mapping multi-modal data into a common subspace defined by label information.
The second term containing $l_{21}$-norms conducts feature selection on two feature spaces ${\bf X}_a$~and~${\bf X}_b$ simultaneously.
The trace norm can enhance the relevance of projected data with connections inside the subspace.

\subsection{Joint Feature Selection and Subspace Learning for Cross-modal Retrieval (JFSSL)}

JFSSL~\cite{DBLP:journals/pami/WangHWWT16} is an extension based on LCFS~\cite{wang2013learning}.
The objective function is a generic minimization problem among $M$ different modalities of data objects in the following form:
\begin{equation}
\label{equ:jfssl}
\begin{split}
\min \limits_{{\bf W}_1, \cdots, {\bf W}_M} \sum_{p=1}^{M} \Vert {\bf X}^T_p {\bf W}_p - {\bf Y} \Vert_F^2 +
\lambda_1 \sum_{p=1}^{M} \Vert {\bf W}_p \Vert_{21} + \lambda_2 \Omega ({\bf W}_1, \cdots, {\bf W}_M)~,
\end{split}
\end{equation}
where ${\bf W}_p~(p = 1, \cdots, M)$ denotes the projection matrix for the $M$-th modality.
The roles of its first term and the second term are the same as those in LCFS.
The third term is a multi-modal graph regularization reinforcing the intra-modality and inter-modality similarity.
Similar to the empirical separability term of CDFE objective,
this multi-modal graph regularization preserves the intra-modality compactness and the inter-modality dispersion.

%These cross-modal subspace learning methods have the common workflow that learning a projection matrix for each modality to project the data from different modalities into a common comparable subspace in the training phase.
%In the testing phase, testing data samples from one modality will be taken as the query set to retrieve matched samples from the other modality.
%\textcolor{blue}{This paragraph is not appeared in the IC-NIDC camera-ready.}

\section{Experimental Results and Discussions}

\subsection{Datasets}

\begin{figure*}[htb]
\centering
\includegraphics[width=0.90\textwidth]{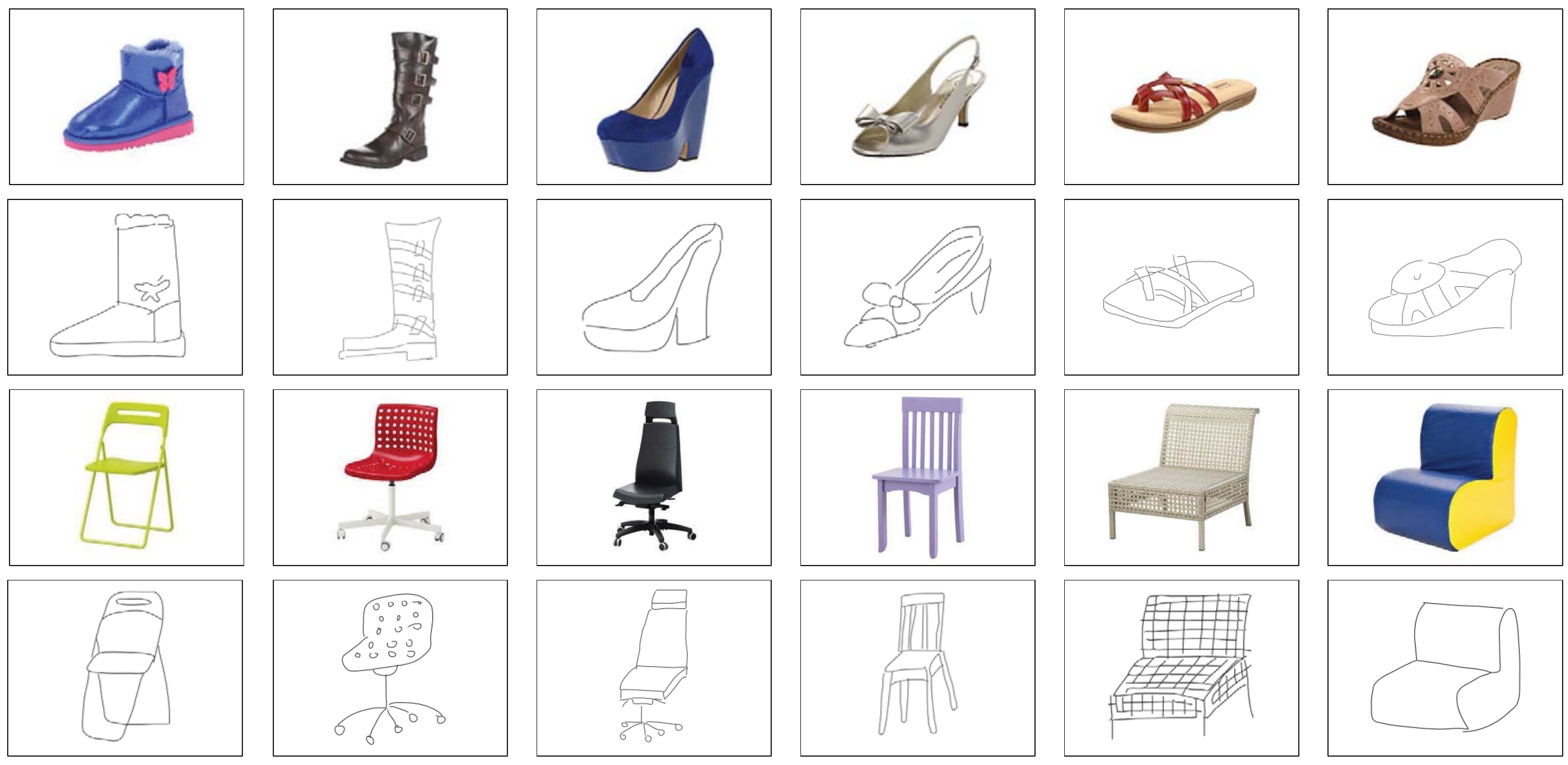}%
\caption{Samples of the shoe dataset and the chair dataset.}
\label{data-samples}
\end{figure*}

In this section, we will apply the aforementioned cross-modal subspace learning methods on two recently released
fine-grained sketch-based image retrieval datasets~\cite{qianyu2016sketch,keli2016fgsbir}.
Each photo has a corresponding freehand sketch.
That is, each sketch sample has a ground-truth photo counterpart as shown in Fig.~\ref{data-samples}.

The shoe dataset has $419$ photo-sketch pairs, which can be categorized into three subclasses.
All the sample pairs are single-labeled.
The chair dataset contains $297$ photo-sketch pairs.
These chairs can be divided into six subclasses.
The sketches are drawn by nonexperts using their fingers on the touch screens, therefore, these sketches are abstract enough to escape the photo modal space.

\subsection{Experimental Settings}

These cross-modal subspace learning methods (i.e., CCA, PLS, BLM, GMLDA, GMMFA, CDFE, CCA-3V, LCFS, JFSSL) were applied to be performed on shoe dataset and chair dataset, for two SBIR retrieval tasks ((1)~photos query sketches and (2)~sketches query photos).
The experimental results contain randomness due to the limitation by the numbers of the samples in the shoe dataset and chair dataset.
To remove the effect of randomness, we repeated each model on each setting $50$ times.
On the shoe dataset, each evaluation we randomly chose $304$ sample pairs as training set,
and treated the remaining $115$ sample pairs as test set.
On the chair dataset, the ratio of training and test data sets was kept as $200$ to $97$.

In the training phase, we input photo and sketch features into these cross-modal subspace learning methods to learn a projection matrix for each modality.
After model training, we used the projection matrices to map the photo and sketch testing samples into a common subspace.
The cosine distance was adopt to measure the similarity between the projected photos and sketches.
Given a photo (or sketch) query, the goal of each SBIR task is to find the nearest neighbors (NN) from the sketch (or photo) database.

In all the following experiments, we used Histogram of Oriented Gradient (HOG) features to describe the photos and sketches.
In order to evaluate the performance of these methods with different scales,
two kinds of metrics were adopted.
The mean average precision (MAP)~\cite{rasiwasia2010new} was applied to evaluate the performances on semi-fine-grained level.
A retrieval was judged as correct by MAP as long as the retrieved sample and the query sample have the same subclass label.
Another metric ``$acc.@K$''~\cite{qianyu2016sketch,keli2016fgsbir} was used to carry out fine-grained evaluation on the instance-level,
which is the percentage of the corresponding photos or sketches ranked in the top $K$ results.

\subsection{Results on Shoe Dataset}

\subsubsection{Evaluation by MAP}

The MAP scores of different cross-modal subspace learning methods on shoe dataset are reported in Table~\ref{table:MAPscores-on-shoe}.
The minimum (min), maximum (max), mean value (mean), variance (var), and standard deviation (std) for each method
are also presented.

Wang et al.~\cite{wang2013learning,DBLP:journals/pami/WangHWWT16} have illustrated that
CCA, PLS, BLM, GMLDA, GMMFA, CDFE, and CCA-3V are incapable of feature selection.
Hence we utilize Principal Component Analysis (PCA) to remove the redundancy in the input features for these seven kinds of methods.
In order to validate the feature selection abilities of LCFS and JFSSL,
their results without performing PCA on the input features are also reported in Table~\ref{table:MAPscores-on-shoe}.
It can be observed that LCFS and JFSSL outperform the remaining methods
for photo querying sketch and sketch querying photo on shoe dataset.
This is because LCFS and JFSSL can simultaneously select discriminative and effective features from different
modalities while learning the common subspace.

In terms of performance, GMLDA and GMMFA are close to LCFS and JFSSL.
The performance gaps between GMLDA, GMMFA, and LCFS, JFSSL are not as obvious as those for image and text matching.
This is due to the inherent data difference between sketch and text.
GMLDA and GMMFA are superior to CDFE and CCA-3V.
Among these cross-modal subspace learning methods, CCA performs the worst
while its supervised enhanced version CCA-3V achieves good performance.
PLM and BLM are a little better than CCA for photo querying sketch and sketch querying photo.

\begin{figure}[h]
	\centering
	\subfigure[\tiny{Original space of $a$ modality}]{
		\label{fig:subfig:comp1}
		\includegraphics[width=0.17\textwidth]{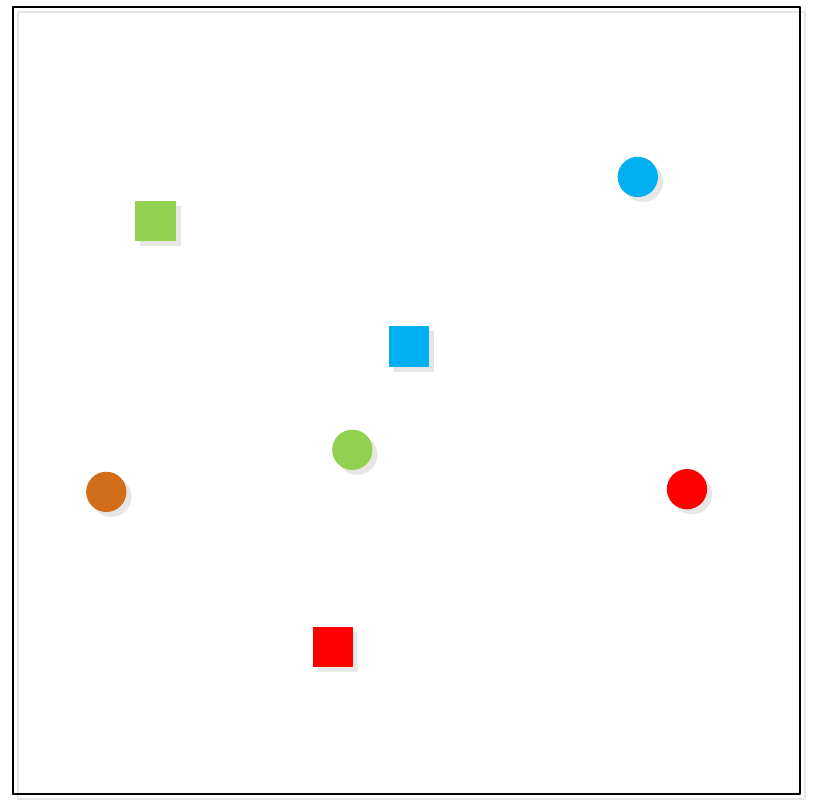}}
	%\hspace{5mm}
	\subfigure[\tiny{Original space of $b$ modality}]{
		\label{fig:subfig:comp2}
		\includegraphics[width=0.17\textwidth]{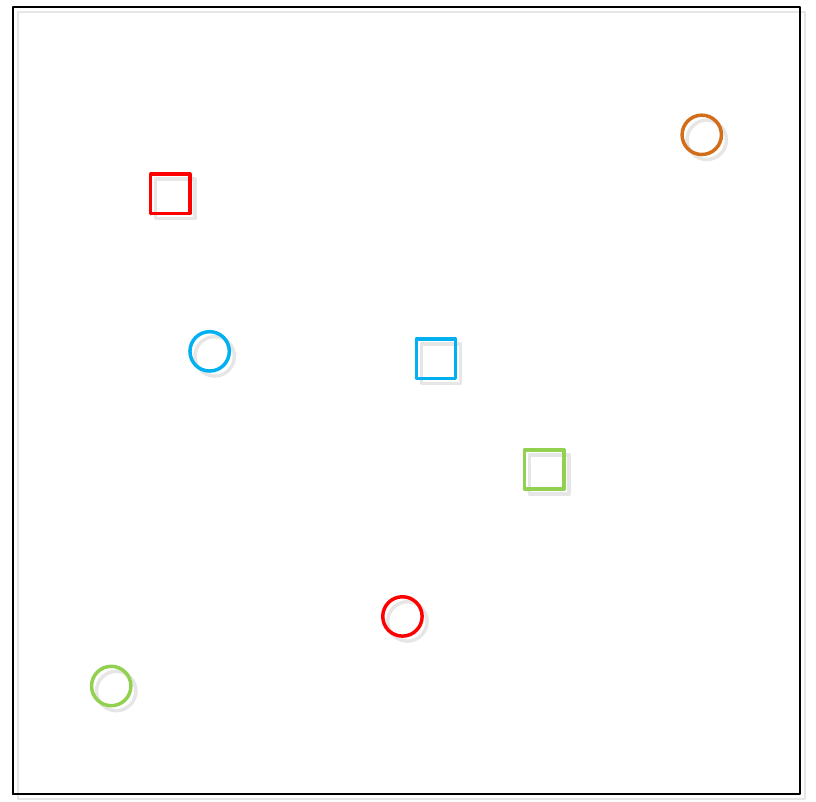}}	
	%\vfill		
	\subfigure[\tiny{Subspace learned by CCA/PLS/BLM}]{
		\label{fig:subfig:comp3}
		\includegraphics[width=0.17\textwidth]{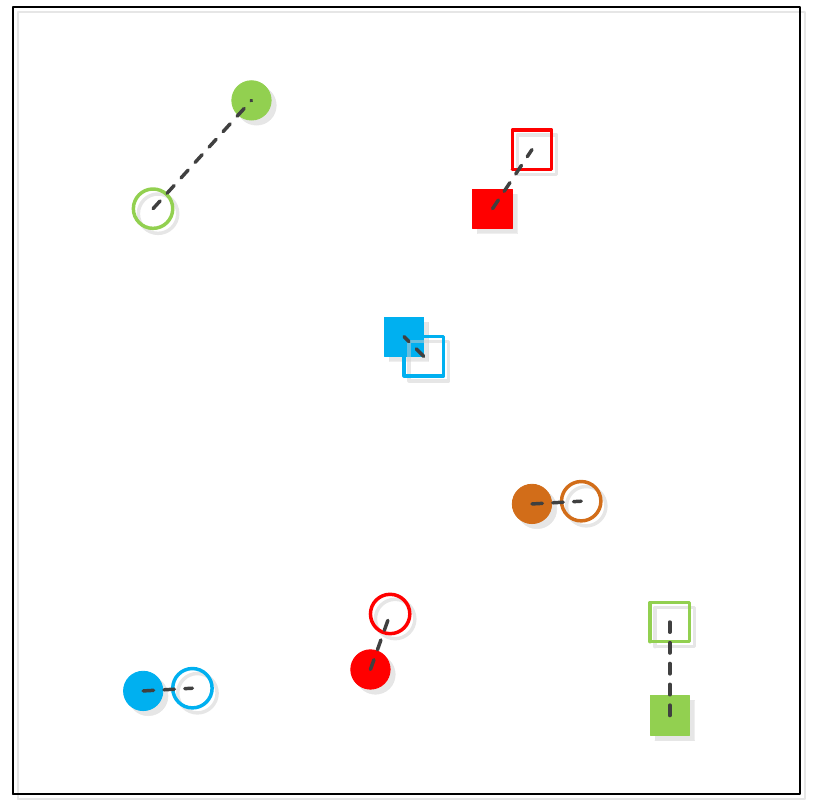}}
	%\hspace{5mm}
	\subfigure[\tiny{Subspace learned by GMLDA/GMMFA/CCA-3V}]{
		\label{fig:subfig:comp4}
		\includegraphics[width=0.17\textwidth]{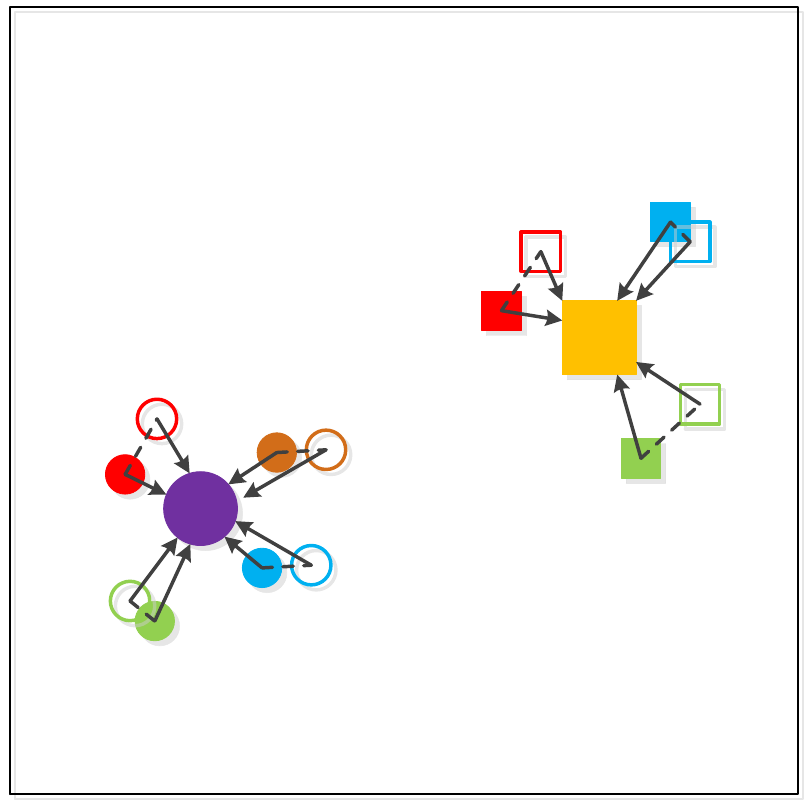}}
    \subfigure[\tiny{Subspace learned by CDFE/LCFS/JFSSL}]{
		\label{fig:subfig:comp5}
		\includegraphics[width=0.17\textwidth]{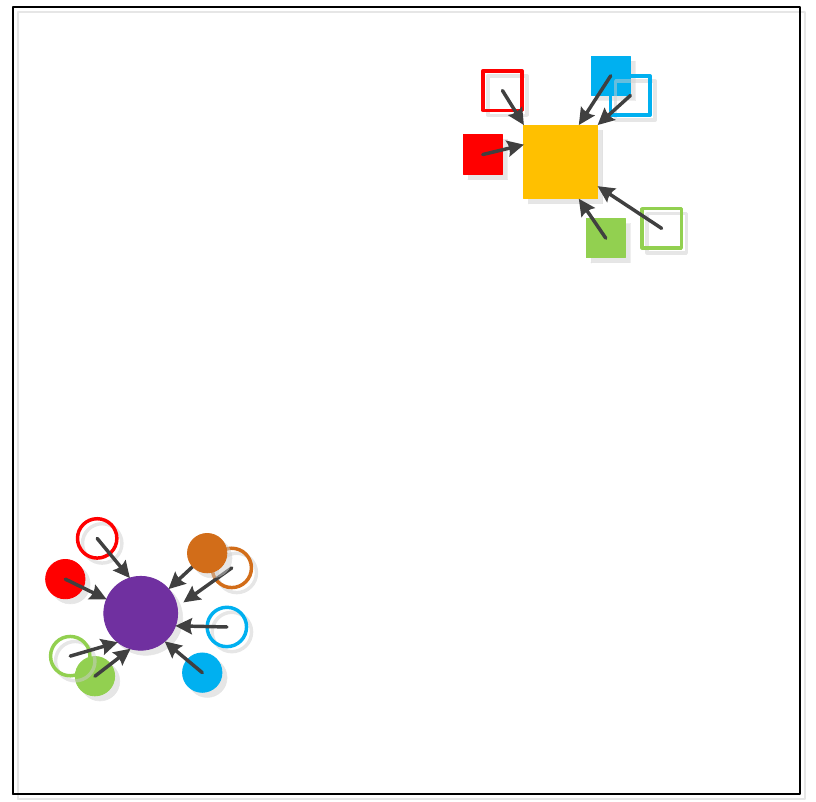}}
	\caption{Demonstration and comparison of various cross-modal approaches.
The smaller solid and hollow figures represent the
data from the different modalities. The smaller figures with the same shape and different colors represents
different samples belonging to the same class. The bigger figures with different shapes indicate the
corresponding domain-independent semantic labels. The short dashed straight lines visualize the pair-wise relationships. The
short bold arrows link multi-modal sample pairs to their semantic labels. The short dashed straight lines
visualize the pair-wise relationships. The short bold arrows link multi-modal sample pairs to their
semantic labels.
}
	\label{fig:comparison}
\end{figure}

The overall trend of Table~\ref{table:MAPscores-on-shoe} can be summarized as supervised methods outperforming the unsupervised methods.
This trend can also be explained by Fig.~\ref{fig:comparison},
which shows the differences between these methods.
CCA, PLS, and BLM used only pair-wise information to build the common subspace, as shown in Fig.~\ref{fig:subfig:comp3}.
Fig.~\ref{fig:subfig:comp4} illustrates that GMLDA, GMMFA, and CCA-3V can take the advantage of class label information and pair-wise relationship to construct preferable inter-class separation in the common subspace.
CDFE mainly attempts to keep the intra-class and inter-class structures in a subspace.
LCFS and JFSSL devote to minimize the subcategory-based residuals.
However, their graph embedding technologies can only improve intra-class compactness and inter-class dispersion.
CDFE, LCFS, and JFSSL cannot thoroughly capture the pair-wise relationship.
In contrast to Fig.~\ref{fig:subfig:comp5},
the sample pairs of Fig.~\ref{fig:subfig:comp4} have dashed lines to connect each other to visualize the pair-wise connections.

These phenomena are consistent with the results presented in~\cite{DBLP:journals/pami/WangHWWT16}.
In~\cite{DBLP:journals/pami/WangHWWT16}, it was discussed and verified that
JFSSL can utilize the multi-modal graph embedding constraint to obtain performance improvements basing on LCFS.
However, for experiments on shoe dataset, their performances are almost the same.
This is because the graph embedding constraint of JFSSL cannot fully play its role on this sketch dataset.

\begin{figure*}[htb]
	\centering
	\subfigure[Photo queries sketch]{
		\label{map-boxplot-photo-query-sketch-on-shoe}
		\includegraphics[width=0.47\textwidth]{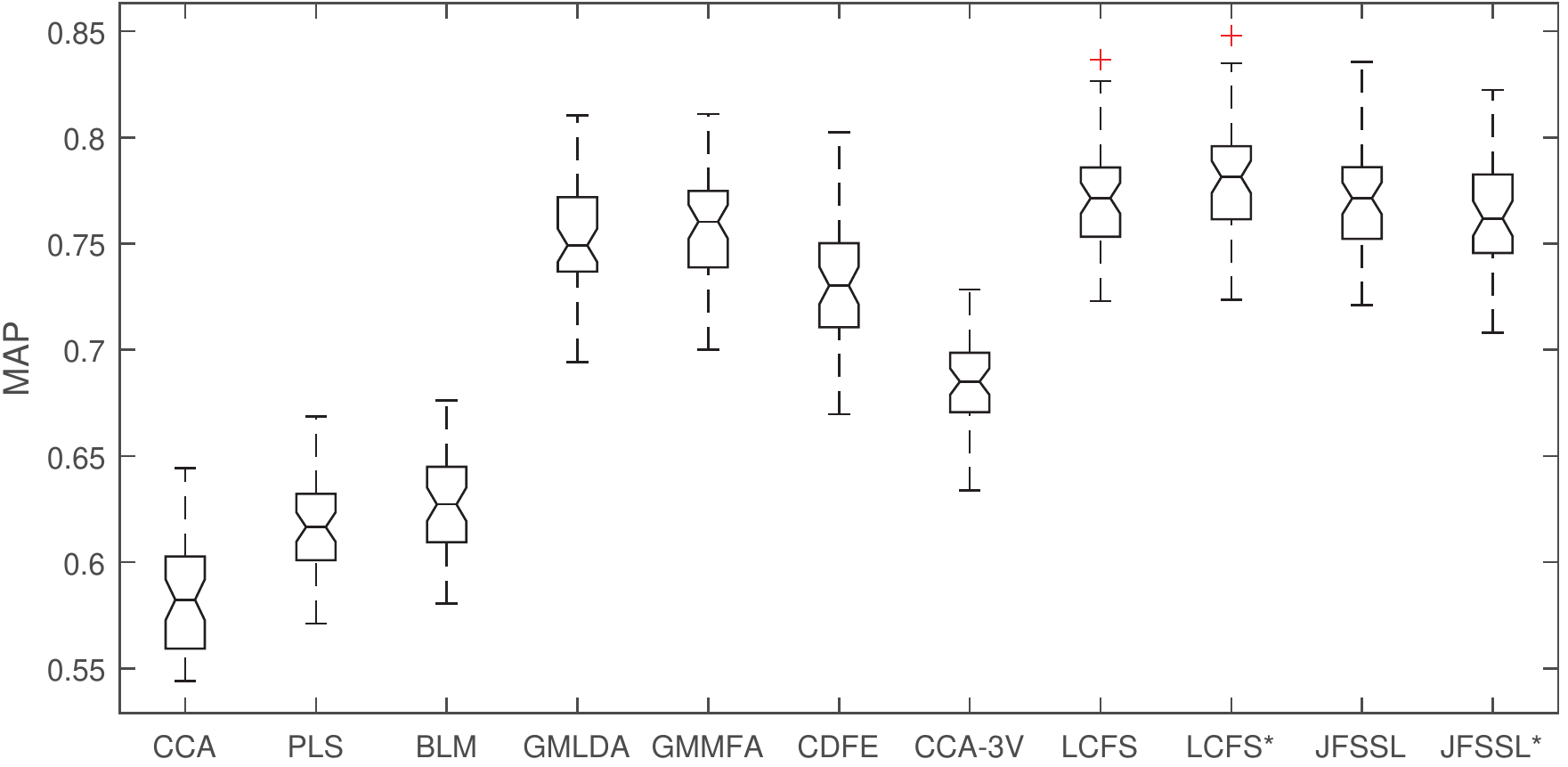}}
	%\hspace{5mm}	
	\subfigure[Sketch queries photo]{
		\label{map-boxplot-sketch-query-photo-on-shoe}
		\includegraphics[width=0.47\textwidth]{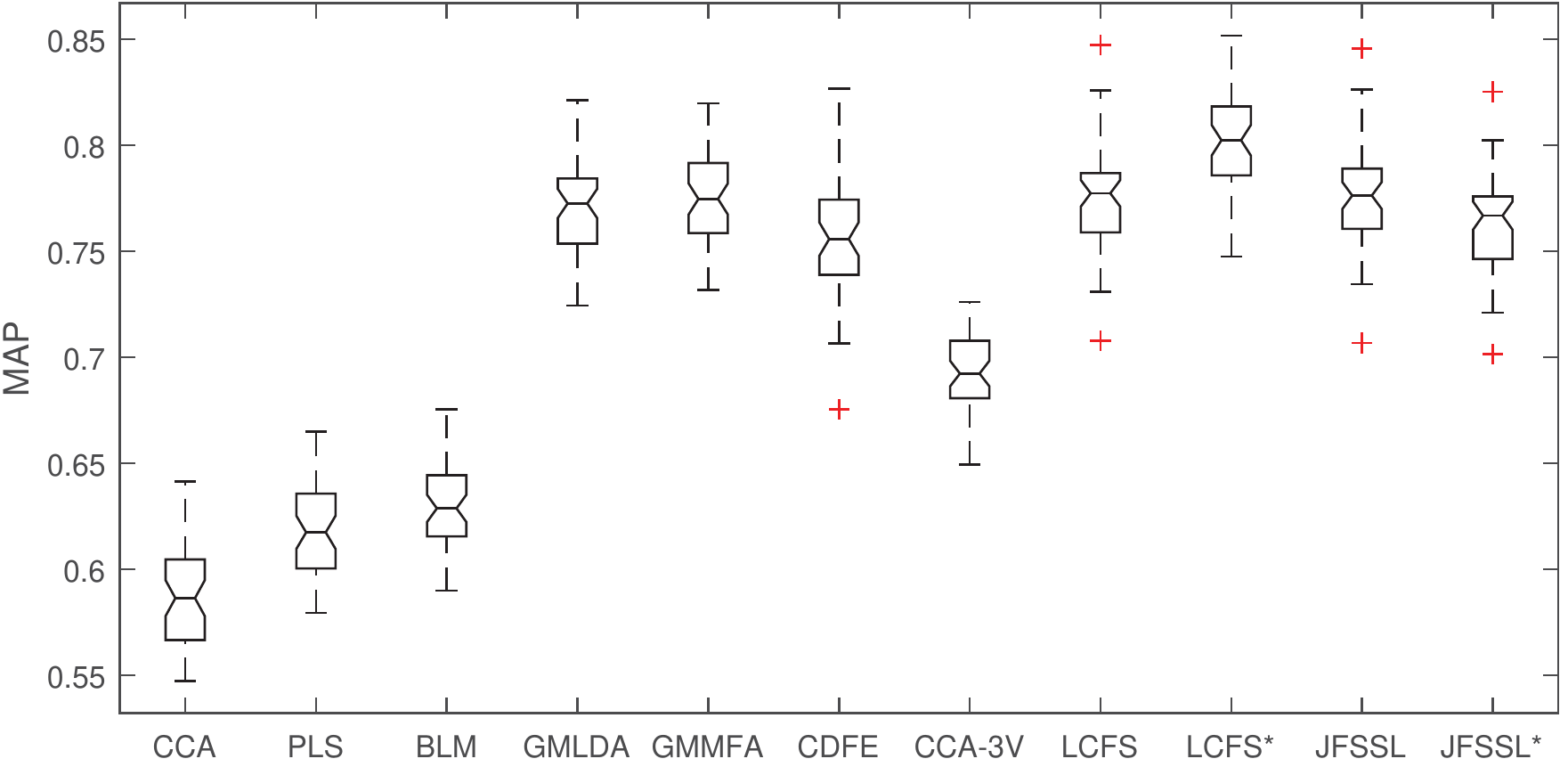}}
	\caption{Box-plots of MAP scores achieved by different cross-modal subspace learning methods on shoe dataset.
The inputs of all the methods are preprocessed by PCA
excepting methods marked with an asterisk.
The top and bottom edges of the box are the 75th and 25th percentiles, respectively.
The outliers are marked as red cross patterns individually.
}
	\label{map-boxplot-on-shoe}
\end{figure*}

All the experimental results in Table~\ref{table:MAPscores-on-shoe} are also visualized as box-plots in Fig.~\ref{map-boxplot-on-shoe}.
The box range of certain method shows the performance stability of corresponding method for the SBIR tasks.
We can conclude that
these cross-modal subspace learning methods have similar stabilities for SBIR on shoe dataset.

All the samples extracted from the same dataset follow the same underlying data distribution.
Each method has its own unique principle and can be regarded as a system.
Theoretically, the experimental results of a method will also follow a certain latent data distribution
when it is repeated on the same dataset.
Thus we can judge that the performances of these aforementioned methods for the SBIR tasks on shoe dataset are fundamentally different,
only when their experimental result distributions do not belong to the same distribution.

According to Table~\ref{table:MAPscores-on-shoe} and Fig.~\ref{map-boxplot-on-shoe},
we get a preliminary conclusion that LCFS is the best among these methods on shoe dataset
for photo querying sketch and sketch querying photo.
To verify whether LCFS is fundamentally superior to other methods,
we conducted student¡¯s t-test between LCFS and other methods,
as shown in Table~\ref{table:ttest-on-shoe}.
The null hypothesis is that the two results have similar means with unknown variance.
We can observe that
LCFS and JFSSL have the same output MAP distribution for photo querying sketch task
no matter whether their input features are preprocessed by PCA.
However, their output MAP distributions for sketch querying photo task are different.
In all other cases, LCFS is statistically different from the others.
Based on the above observations, we conclude that the performance of LCFS
for the subcategory-level SBIR tasks on shoe dataset is essentially different with the performances of
CCA, PLS, BLM, GMLDA, GMMFA, CDFE, and CCA-3V.

\begin{table*}[!t]\scriptsize
\centering

\caption{MAP scores achieved by different cross-modal subspace learning methods on shoe dataset.}
\label{table:MAPscores-on-shoe}
\begin{tabular}{|c|c|c|c|c|c|c|c|c|c|c|}
  \hline \hline

  \multirow{2}{*}{Method} & \multicolumn{5}{|c|}{Photo queries sketch} & \multicolumn{5}{|c|}{Sketch queries photo}  \\
  \cline{2-11}
   %& ~~~min~~~ & ~~~max~~~ & ~~~mean~~~ & ~~~var~~~ & ~~~std~~~ & ~~~min~~~ & ~~~max~~~ & ~~~mean~~~ & ~~~var~~~ & ~~~std~~~ \\
   & ~min~ & ~max~ & ~mean~ & ~var~ & ~std~ & ~min~ & ~max~ & ~mean~ & ~var~ & ~std~ \\
  \hline
     PCA+CCA &0.5442    &0.6442    &0.5836    &0.0007    &0.0264    &0.5474    &0.6415    &0.5868    &0.0006    &0.0255 \\
   PCA+PLS &0.5712    &0.6687    &0.6169    &0.0005    &0.0218    &0.5795    &0.6649    &0.6187    &0.0004    &0.0205 \\
   PCA+BLM &0.5805    &0.6762    &0.6272    &0.0005    &0.0217    &0.5900    &0.6755    &0.6294    &0.0004    &0.0206 \\
   PCA+GMLDA &0.6943    &0.8103    &0.7542    &0.0007    &0.0267    &0.7244    &0.8213    &0.7712    &0.0006    &0.0253 \\
   PCA+GMMFA &0.7000    &0.8111    &0.7577    &0.0006    &0.0248    &0.7317    &0.8199    &0.7733    &0.0005    &0.0227 \\
   PCA+CDFE &0.6696    &0.8024    &0.7302    &0.0008    &0.0277    &0.6755    &0.8268    &0.7559    &0.0007    &0.0271 \\
   PCA+CCA-3V &0.6339    &0.7284    &0.6837    &0.0005    &0.0219    &0.6494    &0.7261    &0.6930    &0.0004    &0.0191 \\
   PCA+LCFS &0.7229    &0.8365    &0.7705    &0.0007    &0.0255    &0.7079    &0.8473    &0.7745    &0.0006    &0.0244 \\
   LCFS &0.7236    &0.8480    & \textbf{0.7798}    &0.0007    &0.0258    &0.7475    &0.8518    & \textbf{0.8014}    &0.0006    &0.0237 \\
   PCA+JFSSL &0.7211    &0.8355    &0.7700    &0.0006    &0.0254    &0.7067    &0.8457    &0.7748    &0.0006    &0.0242 \\
   JFSSL &0.7080    &0.8222    &0.7619    &0.0006    &0.0249    &0.7016    &0.8253    &0.7632    &0.0006    &0.0244 \\
  \hline
\end{tabular}
\end{table*}

\begin{table*}[!t]\scriptsize
\centering

\caption{MAP scores achieved by different cross-modal subspace learning methods on chair dataset.}
\label{table:MAPscores-on-chair}
\begin{tabular}{|c|c|c|c|c|c|c|c|c|c|c|}
  \hline \hline

  \multirow{2}{*}{Method} & \multicolumn{5}{|c|}{Photo queries sketch} & \multicolumn{5}{|c|}{Sketch queries photo}  \\
  \cline{2-11}
   & ~min~ & ~max~ & ~mean~ & ~var~ & ~std~ & ~min~ & ~max~ & ~mean~ & ~var~ & ~std~ \\
  \hline

      PCA+CCA &0.4973    &0.6407    &0.5558    &0.0012    &0.0347    &0.4998    &0.6400    &0.5588    &0.0011    &0.0334 \\
   PCA+PLS &0.5477    &0.6557    &0.5948    &0.0008    &0.0279    &0.5557    &0.6585    &0.5998    &0.0007    &0.0273 \\
   PCA+BLM &0.5435    &0.6549    &0.5942    &0.0007    &0.0260    &0.5541    &0.6507    &0.5987    &0.0006    &0.0241 \\
   PCA+GMLDA &0.6469    &0.7798    &0.7110    &0.0010    &0.0311    &0.6426    &0.7826    &\textbf{0.7077}    &0.0010    &0.0321 \\
   PCA+GMMFA &0.6473    &0.7852    &0.7094    &0.0011    &0.0331    &0.6341    &0.7892    &0.7053    &0.0011    &0.0336 \\
   PCA+CDFE &0.5638    &0.7393    &0.6585    &0.0011    &0.0328    &0.5603    &0.7400    &0.6637    &0.0009    &0.0293 \\
   PCA+CCA-3V &0.5457    &0.6692    &0.6040    &0.0007    &0.0265    &0.5585    &0.6765    &0.6127    &0.0006    &0.0254 \\
   PCA+LCFS  &0.6491    &0.7993    & \textbf{0.7139}    &0.0013    &0.0357    &0.6292    &0.8043    &0.7046    &0.0013    &0.0360 \\
   LCFS &0.6302    &0.7804    &0.7120    &0.0012    &0.0351    &0.6227    &0.7819    &0.7049    &0.0013    &0.0363 \\
   PCA+JFSSL &0.6504    &0.8004    &0.7137    &0.0013    &0.0356    &0.6284    &0.8036    &0.7043    &0.0013    &0.0359 \\
   JFSSL &0.6339    &0.7949    &0.7119    &0.0013    &0.0357    &0.6283    &0.7833    &0.7045    &0.0013    &0.0365 \\
  \hline
\end{tabular}
\end{table*}

\begin{table*}[!t]\scriptsize
\centering
\caption{$P$-value comparisons between LCFS (inputting feature without PCA preprocess) and other cross-modal subspace learning methods on shoe dataset.
The significant level is $0.05$.
}
\label{table:ttest-on-shoe}
\begin{tabular}{|c|c|c|c|c|c|c|c|c|c|c|}
  \hline \hline
   &  PCA+ & PCA+ & PCA+ & PCA+ & PCA+ & PCA+ & PCA+ & PCA+ & PCA+ & \multirow{2}{*}{JFSSL}  \\
   &  CCA & PLS & BLM & GMLDA & GMMFA & CDFE & CCA-3V & LCFS & JFSSL &   \\
  \hline

Photo Query   &5.1e-60    &3.9e-56    &1.2e-53    &4.0e-06    &3.2e-05    &4.8e-15    &1.7e-36    &0.0728    &0.0578   &0.0006\\
Sketch Query  &5.2e-66    &9.4e-64    &3.0e-61    &1.5e-08    &2.8e-08    &2.4e-14    &1.3e-44    &2.0e-07    &2.4e-07    &3.5e-12\\
Average Value &2.5e-60    &1.9e-56    &6.1e-54    &2.0e-06    &1.6e-05    &1.4e-14    &8.9e-37    &0.0364    &0.0289    &0.0003 \\

  \hline
\end{tabular}

\end{table*}

\begin{table*}[!t]\scriptsize
\centering
\caption{
$P$-value comparisons between LCFS (inputting feature without PCA preprocess) and other cross-modal subspace learning methods on chair dataset.
The significant level is $0.05$.
}
\label{table:ttest-on-chair}
\begin{tabular}{|c|c|c|c|c|c|c|c|c|c|c|}
  \hline \hline

   &  PCA+ & PCA+ & PCA+ & PCA+ & PCA+ & PCA+ & PCA+ & PCA+ & PCA+ & \multirow{2}{*}{JFSSL}  \\
   &  CCA & PLS & BLM & GMLDA & GMMFA & CDFE & CCA-3V & LCFS & JFSSL &   \\
  \hline

Photo Query   &2.7e-40    &1.0e-33    &9.9e-35    &0.8698    &0.6971    &4.4e-12    &1.0e-31    &0.7965   &0.8117    &0.9825 \\
Sketch Query  &5.9e-38    &8.8e-30    &2.1e-31    &0.6830    &0.9536    &1.0e-08    &1.4e-26    &0.9633    &0.9309    &0.9516 \\
Average Value &2.9e-38    &4.4e-30    &1.0e-31    &0.7764    &0.8254    &5.3e-09    &7.2e-27    &0.8799    &0.8713    &0.9670 \\

  \hline
\end{tabular}

\end{table*}

\begin{figure*}[htb]
	\centering
	\subfigure[Photo queries sketch]{
		\label{topk-photo-query-sketch-on-shoe}
		\includegraphics[width=0.47\textwidth]{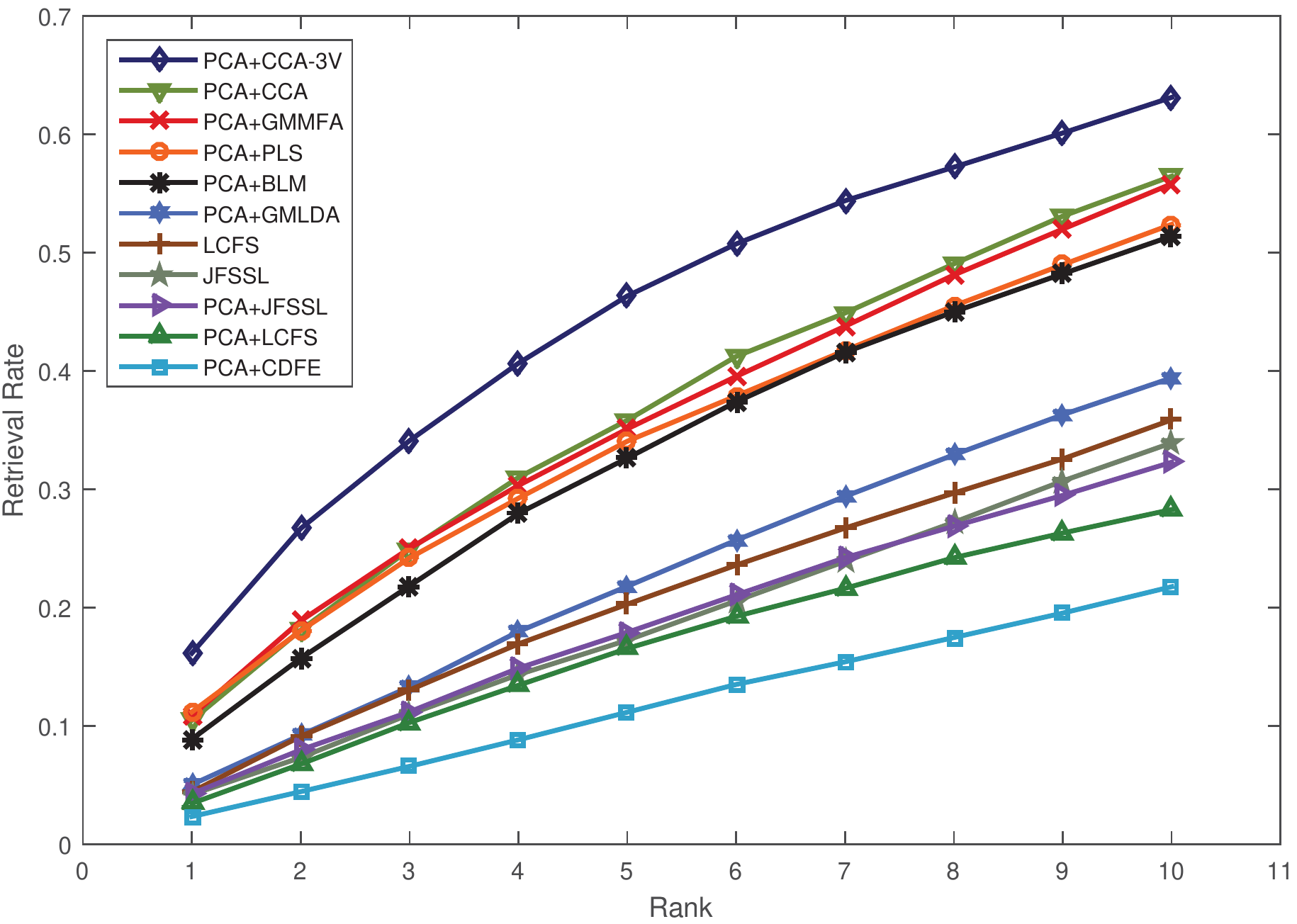}}
	%\hspace{5mm}	
	\subfigure[Sketch queries photo]{
		\label{topk-sketch-query-photo-on-shoe}
		\includegraphics[width=0.47\textwidth]{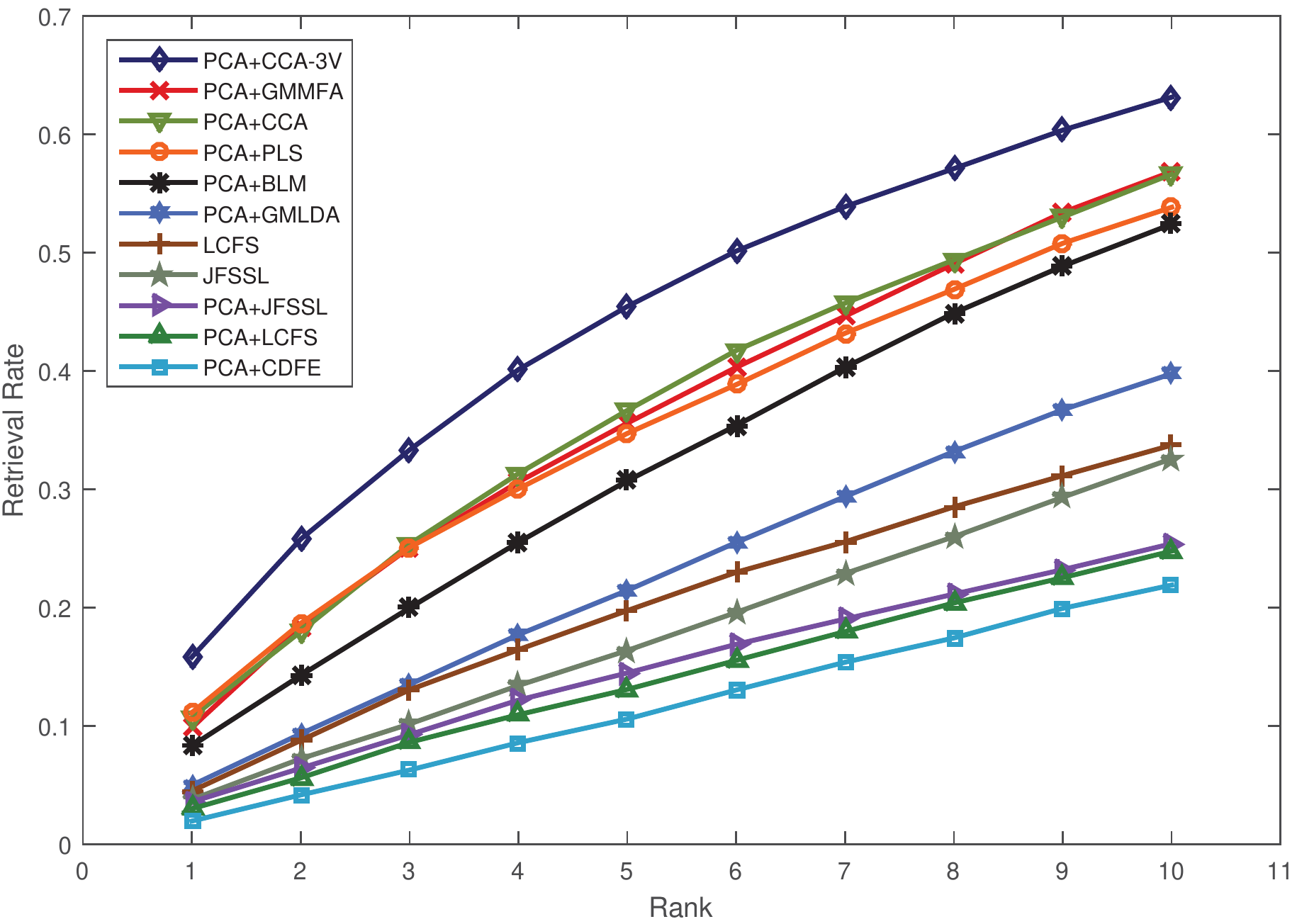}}
	\caption{CMC curves for different cross-modal subspace learning methods on shoe dataset.
}
	\label{topk-on-shoe}
\end{figure*}

\begin{figure*}[htb]
	\centering
	\subfigure[Photo queries sketch]{
		\label{map-boxplot-photo-query-sketch-on-chair}
		\includegraphics[width=0.47\textwidth]{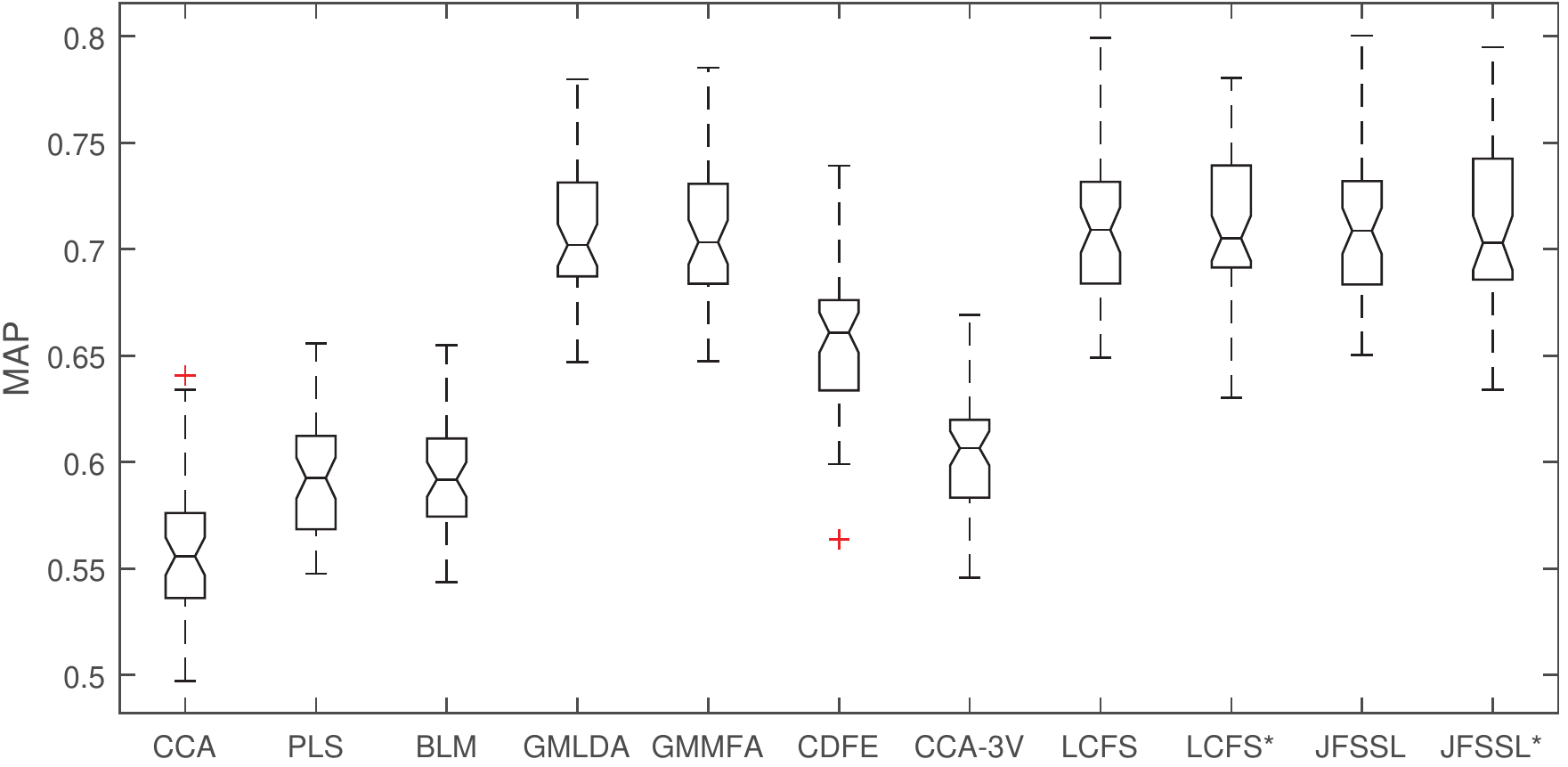}}
	%\hspace{5mm}	
	\subfigure[Sketch queries photo]{
		\label{map-boxplot-sketch-query-photo-on-chair}
		\includegraphics[width=0.47\textwidth]{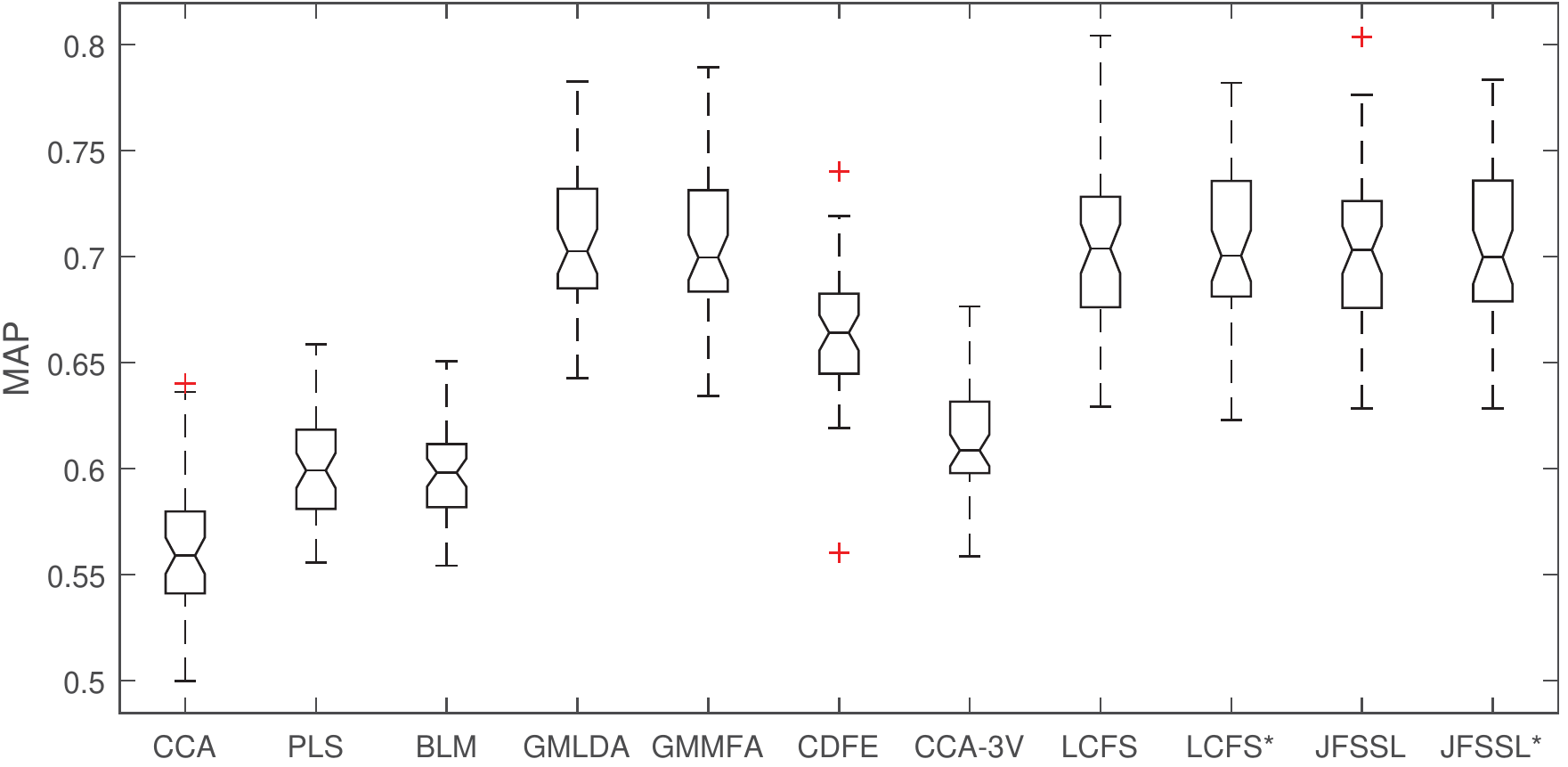}}
	\caption{Box-plots of MAP scores achieved by different cross-modal subspace learning methods on chair dataset.
The inputs of all the methods are preprocessed by PCA
excepting methods marked with an asterisk.
The top and bottom edges of the box are the 75th and 25th percentiles, respectively.
The outliers are marked as red cross patterns individually.
}
	\label{map-boxplot-on-chair}
\end{figure*}

\begin{figure*}[htb]
	\centering
	\subfigure[Photo queries sketch]{
		\label{topk-photo-query-sketch-on-chair}
		\includegraphics[width=0.47\textwidth]{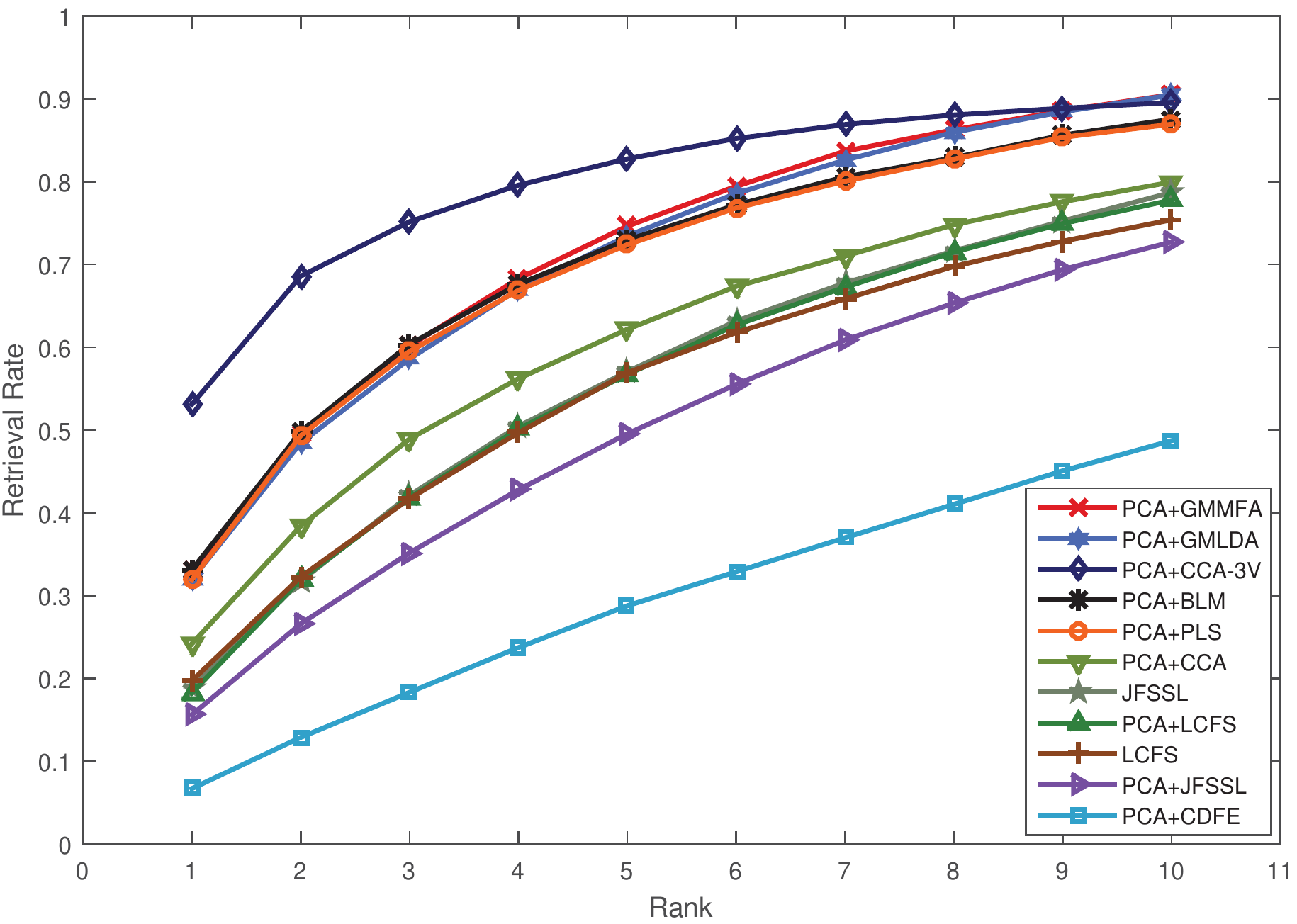}}
	%\hspace{5mm}	
	\subfigure[Sketch queries photo]{
		\label{topk-sketch-query-photo-on-chair}
		\includegraphics[width=0.47\textwidth]{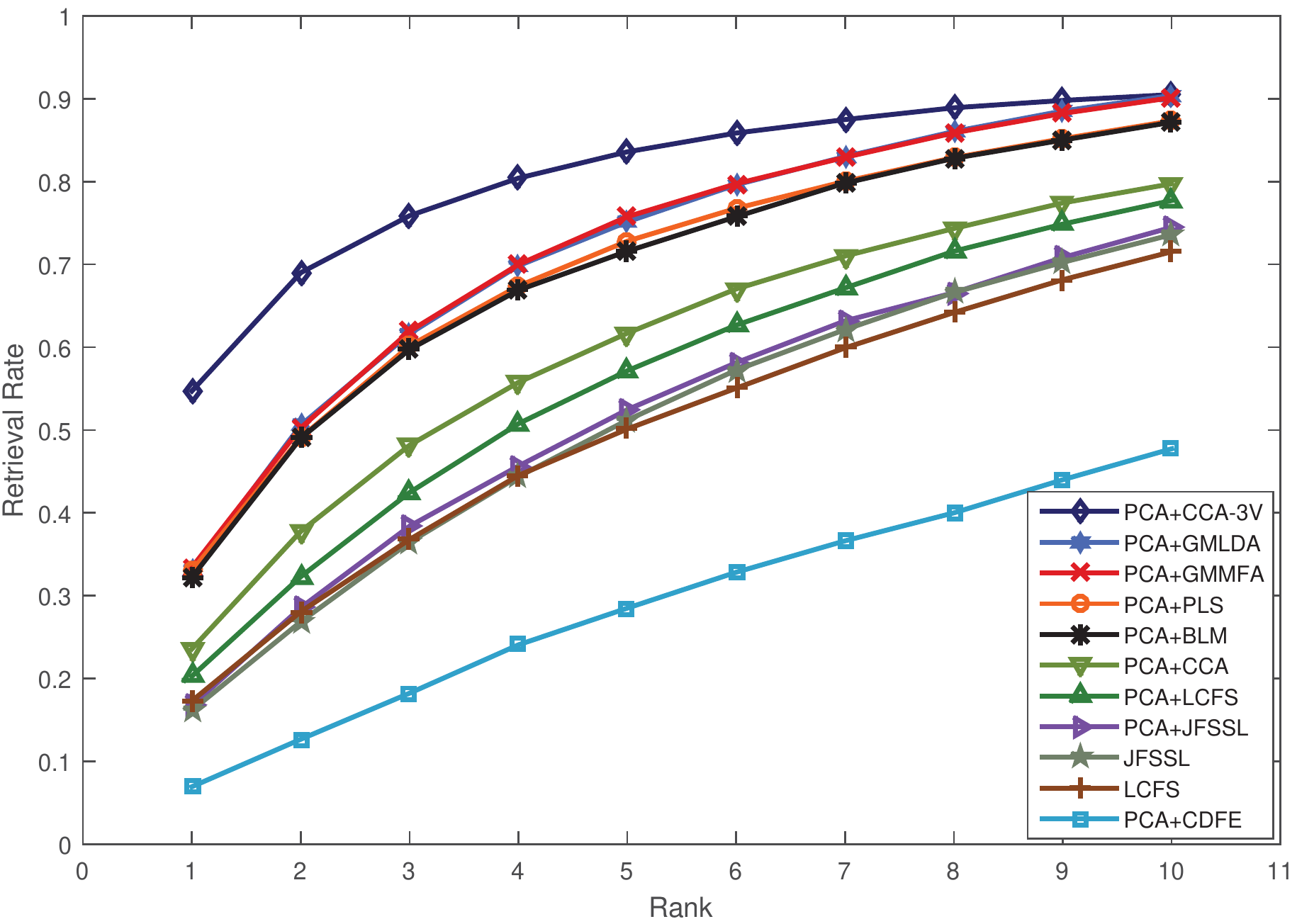}}
	\caption{CMC curves for different cross-modal subspace learning methods on chair dataset.
}
	\label{topk-on-chair}
\end{figure*}

\subsubsection{Evaluation by ``$acc.@K$''}

The shoe dataset and the chair dataset are fine-grained SBIR datasets
in which each sketch sample has a photo sample as its instance-level counterpart.
Hence we can also evaluate the performances of these cross-modal subspace learning methods
by counting the percentage of the corresponding photos or sketches ranked in the top K results.
Please note that the parameters of all the methods are readjusted when we evaluated them by ``$acc.@K$''.

Similar as the previous chapter, PCA is conducted for all the methods.
In addition, to verify the feature selection capabilities of LCFS and JFSSL,
we also input features without PCA reprocessing for these two methods.

The Cumulative Match Characteristic (CMC) curves are plotted in Fig.~\ref{topk-on-shoe}.
We can observe that
in terms of relative distribution relationships and trends of the curves,
Fig.~\ref{topk-photo-query-sketch-on-shoe} is consistent with Fig.~\ref{topk-sketch-query-photo-on-shoe}.
And the performances of these cross-modal subspace learning methods are different from theirs on subcategory-level evaluation (MAP).
CCA-3V achieves the highest instance-level accuracy for photo-sketch query and sketch-photo query on shoe dataset.
The curves of CCA, GMMFA, PLS, and BLM are slightly lower than CCA-3V's.
GMLDA obtains more satisfying experimental results than LCFS and JFSSL.
LCFS and JFSSL are little better than CDFE.
And LCFS and JFSSL still can obviously show their feature selection ability for the instance-level SBIR retrieval on shoe dataset.
The experimental result of CDFE is the worst.

These supervised cross-modal subspace learning methods do not show a distinct advantage over unsupervised methods
for the instance-level SBIR tasks on shoe dataset.
We can conclude that learning pair-wise information is more effective than
learning subcategory-level relationship for instance-level SBIR.
CCA, PLS, and BLM can achieve good results because they can learn the pair-wise relationships of multi-modal samples.
CCA-3V, GMLDA, and GMMFA can utilize sample labels to learn some subcategory separation in the common subspace
in the same time capturing the sample pair-based correlation crossing modalities.
CCA-3V is more focused on modeling the association between the pairs of samples
while GMMFA and GMLDA also learn some structured information in the common subspace.
LCFS and JFSSL cannot obtain the desired results.
For LCFS, its objective function engages in optimizing the subcategory-based residuals and feature selection for each modality.
The trace norm constraint in Eq.~(\ref{equ:lcfs}) can enforce the relevance of projected sample data with connections,
but its weighting coefficient $\lambda_2$ is often too small to learn enough sample pair-wise information.
For JFSSL, its optimization is also mainly minimizing the subcategory-based residuals.
Its graph embedding term in Eq.~(\ref{equ:jfssl}) only preserves the inter-modality and intra-modality similarity.
Hence, LCFS and JFSSL are not good at learning the instance-level or pair-wise relationship of sample data pairs.

\subsection{Results on Chair Dataset}

\subsubsection{Evaluation by MAP}
The MAP score comparison of different cross-modal subspace learning methods on chair dataset
are reported in Table~\ref{table:MAPscores-on-chair}.
Each experiment on each setting is also repeated for 50 times.
And as in the previous chapter, PCA is also utilized to remove the redundancy in the input features
for CCA, PLS, BLM, GMLDA, GMMFA, CDFE, and CCA-3V.
We can observe that
the experimental results are analogous to those on shoe dataset.
Comprehensively considering photo-sketch query and sketch-photo query tasks,
LCFS and JFSSL performs best.
The performances of GMLDA and GMMFA are very close to the performances of LCFS and JFSSL.

The corresponding box-plots for Table~\ref{table:MAPscores-on-chair} are visualized in Fig.~\ref{map-boxplot-on-chair}.
We observe that
the stabilities of these methods for subcategory-level SBIR on chair dataset do not have much difference.
We also conducted the students t-test for the repeated 50 times experimental results between LCFS and other
methods, as shown in Table~\ref{table:ttest-on-chair}.
We observe that GMLDA, GMMFA, LCFS, and JFSSL have the same output
MAP distribution for photo querying sketch and sketch querying photo tasks.

As described above, shoe dataset contains three subcategories and chair dataset has six subcategories.
In common sense, the three-class problem should be easier than the six-class one
when we evaluate the experimental results by MAP.
Thus the MAP score of the same method on shoe dataset should be significantly higher than it on chair dataset.
However, comparing Table~\ref{table:MAPscores-on-shoe} and Table~\ref{table:MAPscores-on-chair},
%the MAP score achieved by some methods
the MAP scores in Table~\ref{table:MAPscores-on-shoe} are not obviously higher than
their counterpart values in Table~\ref{table:MAPscores-on-chair}.
Moreover, Fig.~\ref{map-boxplot-on-shoe} has many outliers (marked red)
while Fig.~\ref{map-boxplot-on-chair} shows almost no outliers.
These phenomena can be interpreted as that
these shoe sketches are not drew very well.
Shoe dataset is mixed with too much noise due to that those shoe sketch samples were painted too rough.

\subsubsection{Evaluation by ``$acc.@K$''}

We readjust the parameters for all the methods and
compare their performances by counting the percentage
of the corresponding photos or sketches ranked in the top K results.
The CMC curves are plotted in Fig.~\ref{topk-on-chair}.
For photo querying sketch task and sketch querying photo task, CCA-3V outperforms other methods.
And the curves of GMMFA, GMLDA, BLM, and PLS in Fig.~\ref{topk-photo-query-sketch-on-chair}
and Fig.~\ref{topk-sketch-query-photo-on-chair} almost overlap together respectively.
We can observe that in Fig.~\ref{topk-photo-query-sketch-on-chair},
`LCFS' curve is slightly lower than `PCA+LCFS' curve
and `JFSSL' curve locates at a high distance above `PCA+JFSSL' curve.
In Fig.~\ref{topk-sketch-query-photo-on-chair}, `LCFS' curve is significantly lower than `PCA+LCFS' curve
and `JFSSL' curve and `PCA+JFSSL' curve are overlapped.
This illustrates that the feature selection abilities of LCFS and JFSSL cannot work well
for instance-level SBIR on chair dataset.
In the objective functions of LCFS and JFSSL, the constraint terms for feature selection are optimized with
the subcategory-based regression residuals simultaneously.
Thus the effect of their feature selection is to reduce the subcategory-based errors rather than instance-level matching errors.

\subsection{Feature Selection and Graph Embedding}

\begin{figure}[h]
	\centering
	\subfigure[Photo queries sketch on shoe dataset]{
		\label{shoe_jfssl_map1}
		\includegraphics[width=0.45\textwidth]{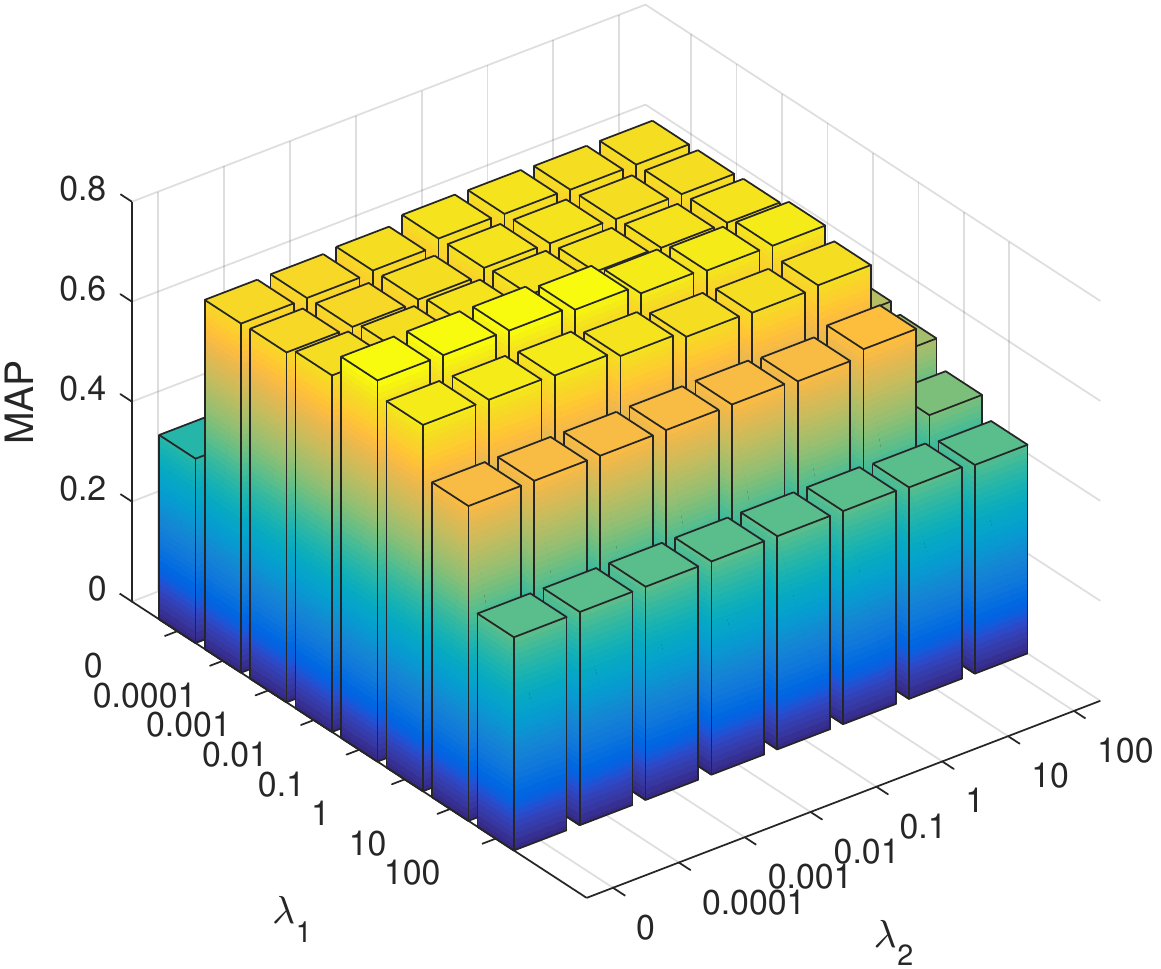}}
	%\hspace{5mm}
	\subfigure[Sketch queries photo on shoe dataset]{
		\label{shoe_jfssl_map2}
		\includegraphics[width=0.45\textwidth]{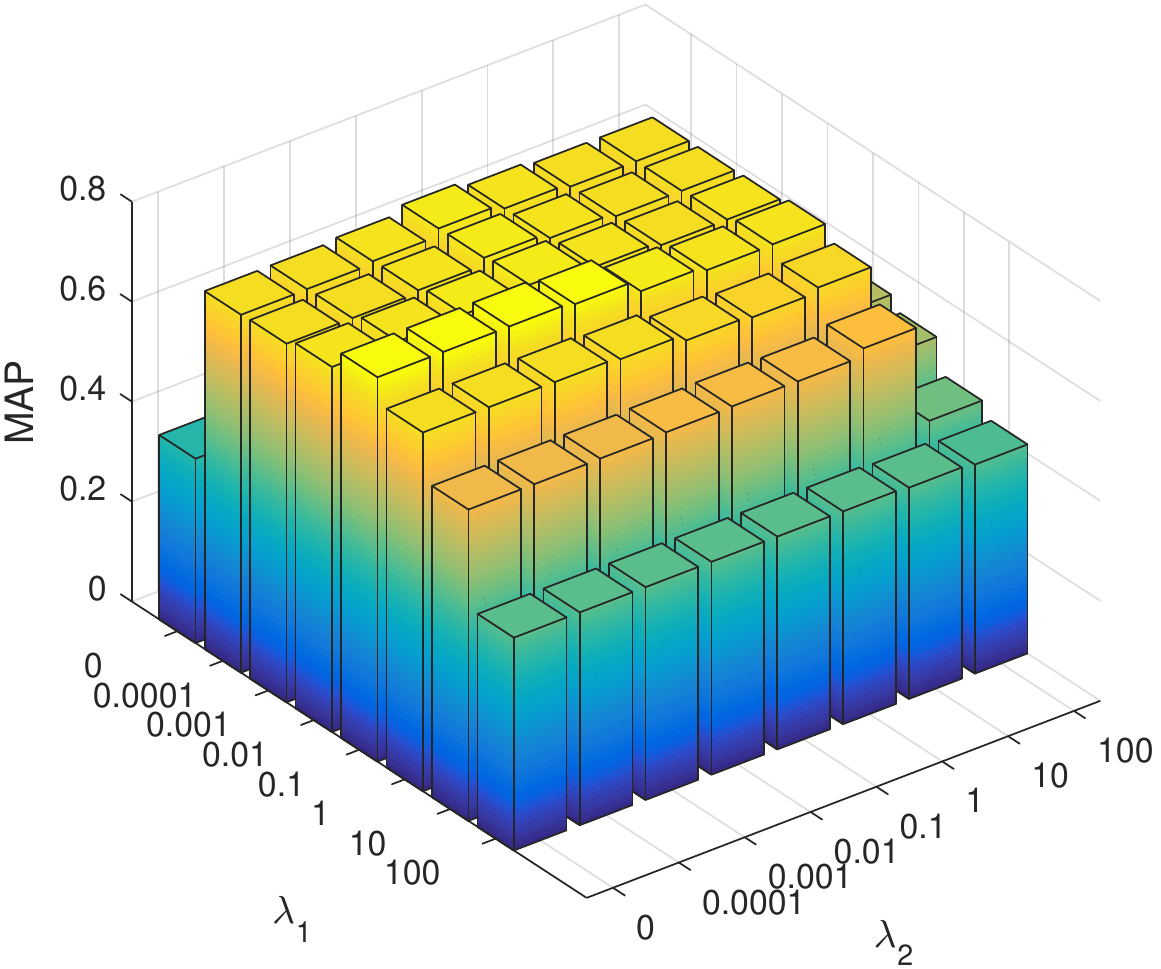}}	
	\vfill		
	\subfigure[Photo queries sketch on chair dataset]{
		\label{chair_jfssl_map1}
		\includegraphics[width=0.45\textwidth]{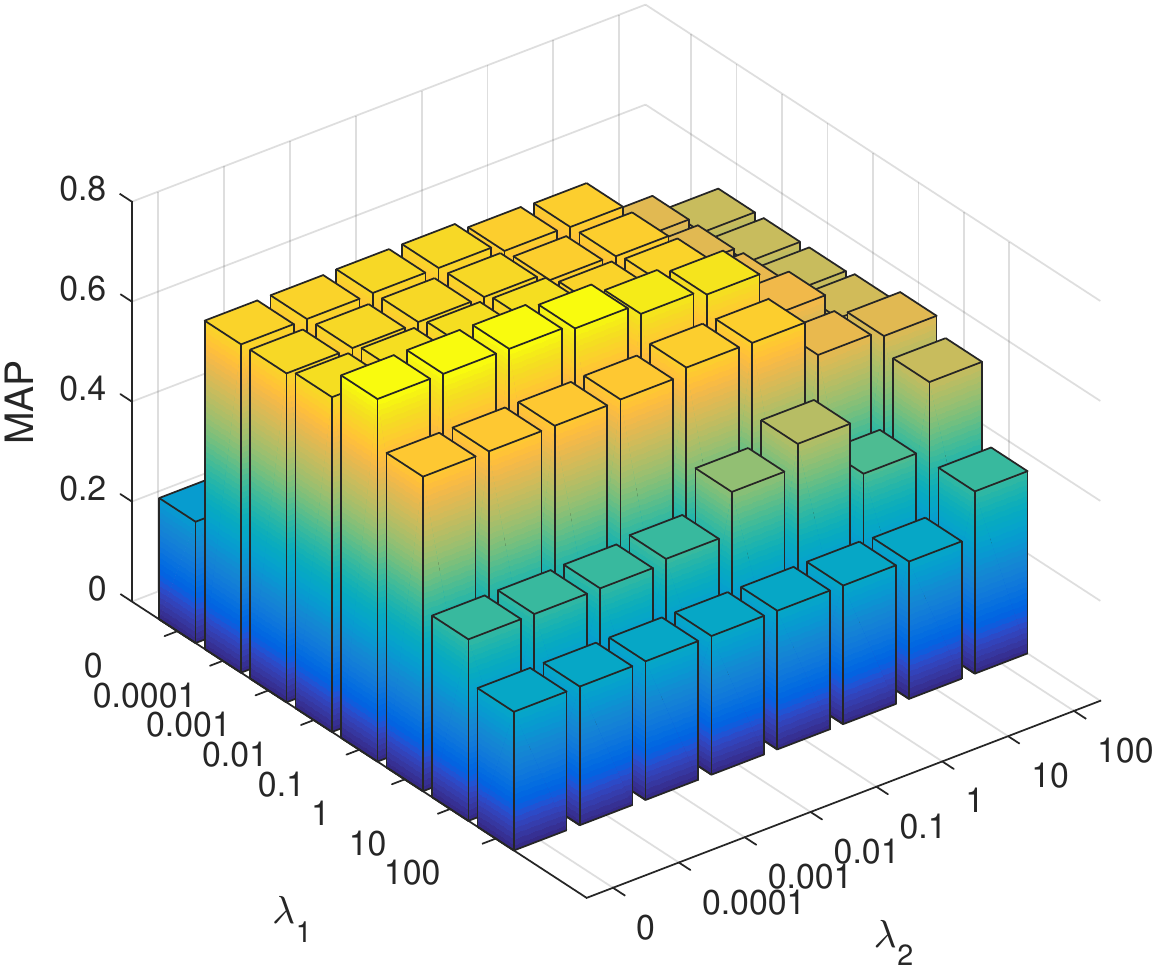}}
	%\hspace{5mm}
	\subfigure[Sketch queries photo on chair dataset]{
		\label{chair_jfssl_map2}
		\includegraphics[width=0.45\textwidth]{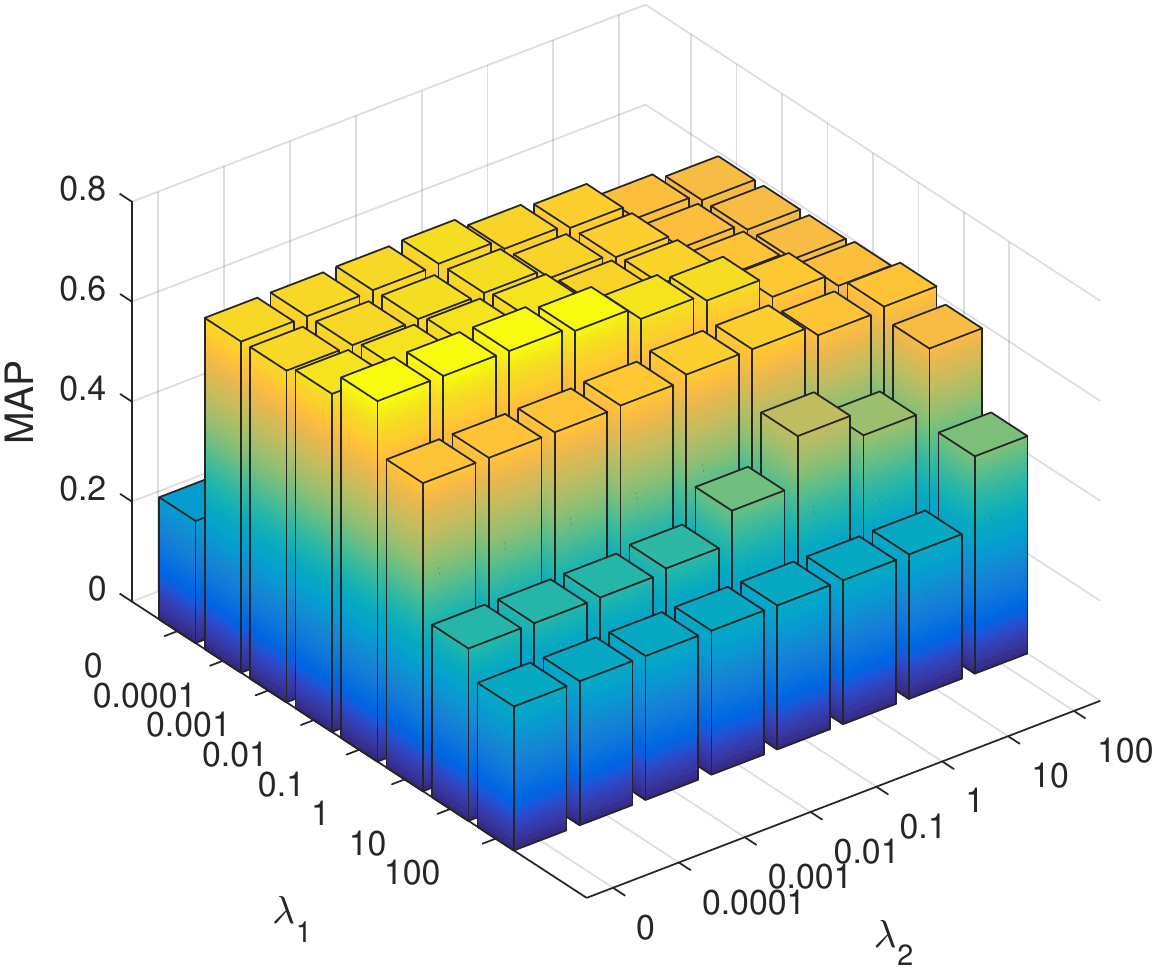}}
	\caption{JFSSL MAP variation with respect to $\lambda_1$ and $\lambda_2$ while its remaining parameters are fixed.}
	\label{fig:JFSSL-adjusting-parameters}
\end{figure}

In the experiments of this paper, the performances of LCFS and JFSSL are almost the same on shoe dataset and chair dataset.
However, JFSSL is the improved version of LCFS and owns theoretical advantages.
JFSSL has feature selection constraint and graph embedding constraint
which are classical operational processes or technologies for cross-modal matching.
Hence, it is worth exploring the synergy between the feature selection and graph embedding terms for SBIR tasks.
In its objective function Eq.~(\ref{equ:jfssl}), $\lambda_1$ and $\lambda_2$ are the weighting parameters
for feature selection and graph embedding terms, respectively.
We tune $\lambda_1$ and $\lambda_2$ in the range of
$\{0, 0.0001, 0.001, 0.01, 0.1, 1, 10, 100\}$ fixing the remaining parameters.
This adjusting process is illustrated in Fig.~\ref{fig:JFSSL-adjusting-parameters}.
We can observe a smooth and symmetric correlation variation between $\lambda_1$ and $\lambda_2$.
This shows us that these two techniques can co-work harmoniously for SBIR.
When $\lambda_1$ is fixed, MAP value slightly changes with the variations of $\lambda_2$.
MAP varies with $\lambda_1$ while $\lambda_2$ is set to a certain value.
This proves that the performance of JFSSL is largely determined by its feature selection technology.
The importances of the feature selection technology and the graph embedding technology are not equal
in the optimization process of JFSSL for subcategory-level SBIR.
This inspires us to further explore these two techniques in our future research for sketch.

\subsection{Complexity Analysis}

In this section, the computational complexity of each compared cross-modal subspace learning method is discussed briefly.
The asymptotic time complexity of CCA is $\mathcal{O}(d^3)$~\cite{rasiwasia2014cluster} where $d = \max(d_a, d_b)$.
PLS is a fitting model embedding regression technique, for which its complexity is defined in terms of its Degrees of Freedom~\cite{RePEc:bes:jnlasa:v:106:i:494:y:2011:p:697-705}.
GMA can be formulated as a standard generalized eigenvalue problem and solved by any eigenvalue solving technique~\cite{sharma2012generalized}.
CDFE can be solved using an alternate optimization strategy including a main step that is a convex quadratic optimization program with linear constraint~\cite{lin2006inter}.
The approximate kernel maps can be adopted to solving CCA-3V~\cite{gong2014multi} reducing the size of this problem to $(\tilde{d}_1+\tilde{d}_2+\tilde{d}_3)\times (\tilde{d}_1+\tilde{d}_2+\tilde{d}_3)$, where $\tilde{d}_i~(i = 1, 2, 3)$ are the dimensionalities of the respective explicit mappings.
The complexity of LCFS is $\mathcal{O}(d^3+n^{2.376})$~\cite{wang2013learning} where $d = \max(d_a, d_b)$.
The complexity of JFSSL can be denoted as $\mathcal{O}(dn^2+d^2)$~\cite{DBLP:journals/pami/WangHWWT16} where $d = \max(d_a, d_b)$.

For rigorous comparison, the running time for learning the projection matrices is compared among these cross-modal subspace learning methods.
Each methods on each setting are repeated 50 times.
The average running time are reported in Table~\ref{table:Running-time} which reveals that the feature selection processing is time-consuming.
All the MATLAB codes are run on a 2.40GHz server with 64G RAM.

\begin{table*}[!t]\scriptsize
\centering
\caption{
Running time comparison in the unit of second.
}
\label{table:Running-time}
\begin{tabular}{|c|c|c|c|c|c|c|c|c|c|c|c|}
  \hline \hline

     &PCA+ & PCA+ & PCA+ & PCA+ & PCA+ & PCA+ & PCA+ & PCA+ & \multirow{2}{*}{LCFS} & PCA+ & \multirow{2}{*}{JFSSL}  \\
     &CCA & PLS & BLM & GMLDA & GMMFA & CDFE & CCA-3V & LCFS &  & JFSSL &   \\
  \hline
     shoe &0.793    &4.217    &5.609    &5.497    &10.287    &5.558    &4.035    &5.124   &139.665    &5.357    &160.894 \\
     chair&0.691    &7.189    &5.979    &7.454    &7.416    &5.200    &2.494    &2.785   &122.104    &2.938    &137.910 \\

  \hline
\end{tabular}

\end{table*}

\subsection{Discussion}

Our experimental results demonstrate that the cross-modal subspace learning methods designed for image and text can be applied in subcategory-level and instance-level SBIR tasks.
The main advantage of cross-modal subspace learning for SBIR is its clear physical significance.
Their performance rankings for subcategory-level SBIR tasks are almost consistent with those in cross-modal retrieval for image and text.
For subcategory-level SBIR, the class label information is useful and supervised methods are usually superior to unsupervised methods.
Feature selection and graph embedding technologies are also efficient to subcategory-level SBIR and they can work together well.
Their performance rankings for instance-level SBIR tasks are not the same as those for subcategory-level retrieval.
Learning pair-wise information is more effective than learning subcategory-level relationship for instance-level SBIR.
Supervised learning has no significant advantages over unsupervised methods for instance-level SBIR task.
On the shoe dataset and the chair dataset, LCFS outperforms other methods for subcategory-level SBIR and CCA-3V achieves the highest accuracy for instance-level SBIR.
This leads us to conclude that subcategory-level information can also be beneficial to instance-level SBIR.

\section{Discussion and Future work}

We have demonstrated the feasibility of utilizing cross-modal subspace learning methods to tackle the domain-gap between sketches and photos.
In the future, we may gain access to better solutions for SBIR by including the advantages of the cross-modal subspace learning techniques, e.g.,
pair-wise modeling, subcategory-based residual, joint feature selection, graph embedding.
In particular, many researchers use deep Convolutional Neural Network (CNN) to conduct cross-modal matching~\cite{wang2015mmss,xiong2015conditional,Yan:2016:CVS:2964284.2967252} or SBIR
which is essentially to learn some feature subspaces to match multi-modal data.
Moreover, the convolutional sparse coding technology can also learn subspace satisfying certain qualities~\cite{7045929,gwon2016multimodal,zhu2015convolutional}, which illustrates the convolutional idea and subspace learning can be reasonably combined.
Therefore, it is natural to also utilize cross-modal subspace learning concepts to improve CNN for SBIR, and potentially incorporating saliency information~\cite{7436783, 7514760} to improve part-level examination in the same network.

If we assume that sketch sits between photo and text in terms of their expressive power, i.e., photo is the most expressive for it can capture a like-for-like depiction of the visual world, sketches are unlikely to do so since they are highly abstract yet still visual, text on the other hand can be vague and more importantly not in the visual domain anymore.
This bears the question that if modeling sketch together with text and photo could be worthwhile to better bridge the gap between text and photo, e.g., for text-based image retrieval. The fact that CCA-3V achieved the best performance for the fine-grained case is a good indicator of the promise that such three-way modelling offers. However, currently available SBIR datasets cannot provide detailed and adequate semantic textual information. Hence new datasets that capture all three domains are required.

\section{Conclusion}

In this paper, we discussed and evaluated a series of state-of-the-art cross-modal subspace learning methods.
We described each method
and applied these approaches to two recently released fine-grained SBIR datasets.
This paper provided detailed comparisons and analysis on experimental results
and discussed future research opportunities for SBIR.

\section{Acknowledgement}
This work was partly supported by National Natural Science Foundation of China (NSFC) grant No.~$61402047$, NSFC-RS joint funding grant
Nos.~$61511130081$ and IE$141387$, Beijing Natural Science Foundation (BNSF) grant No.~$4162044$, Beijing Nova Program Grant 2017045, the Open Projects Program of National Laboratory of Pattern Recognition grant
No.~$201600018$, and Chinese $111$ program of Advanced Intelligence and Network Service under
grant No. B$08004$.
This work is partly supported by BUPT-SICE Excellent Graduate Students Innovation Fund $(2016)$.

\bibliographystyle{model1-num-names}
\bibliography{egbib}

\end{document}